\pdfoutput=1
\documentclass[10pt,twocolumn,letterpaper]{article}

\usepackage[pagenumbers]{cvpr} 

\usepackage{graphicx}
\usepackage{amsmath}
\usepackage{amssymb}
\usepackage{booktabs}
\usepackage{fancyhdr}
\usepackage[accsupp]{axessibility}

\usepackage[pagebackref,breaklinks,colorlinks]{hyperref}

\usepackage[capitalize]{cleveref}
\crefname{section}{Sec.}{Secs.}
\Crefname{section}{Section}{Sections}
\Crefname{table}{Table}{Tables}
\crefname{table}{Tab.}{Tabs.}

\def\cvprPaperID{6367} 
\def\confName{CVPR}
\def\confYear{2022}

\usepackage{overpic}
\usepackage{enumitem} 
\usepackage{overpic} 
\usepackage{color}
\usepackage{algorithm}
\usepackage{algorithmicx}
\usepackage{algpseudocode}
\usepackage{graphicx}
\usepackage{comment}

\definecolor{turquoise}{cmyk}{0.65,0,0.1,0.3}
\definecolor{purple}{rgb}{0.65,0,0.65}
\definecolor{dark_green}{rgb}{0, 0.5, 0}
\definecolor{orange}{rgb}{0.8, 0.6, 0.2}
\definecolor{red}{rgb}{0.8, 0.2, 0.2}
\definecolor{darkred}{rgb}{0.6, 0.1, 0.05}
\definecolor{blueish}{rgb}{0.0, 0.3, .6}
\definecolor{light_gray}{rgb}{0.7, 0.7, .7}
\definecolor{pink}{rgb}{1, 0, 1}
\definecolor{greyblue}{rgb}{0.25, 0.25, 1}

\usepackage{blindtext}

\renewcommand{\paragraph}[1]{\vspace{1em}\noindent\textbf{#1}.}
\begin{document}
\title{TO-FLOW: Efficient Continuous Normalizing Flows with Temporal Optimization adjoint with Moving Speed}

\author{Shian Du$^{1}$\thanks{Equal contribution. Order determined by coin toss.}, Yihong Luo$^{1,2*}$, Wei Chen$^{1*}$, Jian Xu$^{1}$, Delu Zeng$^{1}$\thanks{Corresponding author: Delu Zeng.}~\hspace{1pt}   \\
	$^1$ South China University of Technology  \hspace{1pt}
    $^2$ Hong Kong University of Science and Technology \\
    \tt\small{201930230264@mail.scut.edu.cn},
    \tt\small  {yihongluo@ust.hk},
    \tt\small {\{202120130414,202010106028\}@mail.scut.edu.cn},
    \\
    \tt\small {dlzeng@scut.edu.cn}
 }
\maketitle
\begin{abstract}
Continuous normalizing flows (CNFs) construct invertible mappings between an arbitrary complex distribution and an isotropic Gaussian distribution using Neural Ordinary Differential Equations (neural ODEs). It has not been tractable on large datasets due to the incremental complexity of the neural ODE training. Optimal Transport theory has been applied to regularize the dynamics of the ODE to speed up training in recent works. In this paper, a temporal optimization is proposed by optimizing the evolutionary time for forward propagation of the neural ODE  training. In this appoach, we optimize the network weights of the CNF alternately with evolutionary time by coordinate descent. Further with temporal regularization, stability of the evolution is ensured. This approach can be used in conjunction with the original regularization approach. We have experimentally demonstrated that the proposed approach can significantly accelerate training without sacrifying performance over baseline models.


\end{abstract}
\section{Introduction}
\label{sec:intro}
As an impressive example of unsupervised learning, deep generative models have exhibited powerful modeling performance across a wide range of tasks, including variational autoencoders (VAE) \cite{kingma2014stochastic}, Generative Adversarial Nets (GAN) \cite{goodfellow2014generative}, autoregressive models \cite{bahdanau2015task} and Flow-based models \cite{dinh2016density,kingma2018glow}.

Generative models based on normalizing flows have been recently realized with great success in the problems of probabilistic modeling and inference. Normalizing flows \cite{rezende2015variational} provide a general and extensible framework for modelling highly complex and multimodal distributions through a series of differentiable and invertible transformations. Since these transformations are invertible, the framework of normalizing flows allows for powerful exact density estimation by computing the Jacobian determinant \cite{ccinlar2013real}. Like a fluid flowing through a set of tubes, the initial density ‘flows’ through the sequence of invertible mappings by repeatedly applying the rule for change of variables until a desired probability at the end of this sequence is obtained. Normalizing flows are an increasingly active area of machine learning research. Applications include image generation \cite{kingma2018glow,ho2019flow++}, noise modelling \cite{abdelhamed2019noise}, video generation \cite{kumar2019videoflow}, audio generation \cite{esling2019universal,kim2018flowavenet,prenger2019waveglow}, graph generation \cite{madhawa2019graphnvp}, reinforcement learning \cite{mazoure2020leveraging,ward2019improving,touati2020randomized}, computer graphics \cite{muller2019neural}, and physics \cite{rezende2020normalizing,kohler2020equivariant,satorras2021n,wirnsberger2020targeted,wong2020gravitational}. A key strength of normalizing flows is their expressive power as generative models due to its ability to approximate the posterior distribution arbitrarily, while maintaining explicit parametric forms. 

Thanks to recent advances in deep generative architectures using maximum likelihood estimation and approximate approach like VAE for large-scale probabilistic models, continuous normalizing flows (CNF) obtained by solving an ordinary differential equations (ODE) were later developed in neural ODEs \cite{chen2018neural}. Neural ODEs form a family of models that approximate the ResNet architecture by using continuous-time ODEs. The neural ODE’s dynamics can be chosen almost arbitrarily while ensuring invertibility. The jump to continuous-time dynamics affords a few computational benefits over its discrete-time counterpart, namely the presence of a trace in place of a determinant in the evolution formulae for the density, as well as the adjoint method for memory-efficient backpropagation. Due to their desirable properties, such as invertibility and parameter efficiency, neural ODEs have attracted increasing attention recently. For example, Grathwohl et al \cite{grathwohl2018ffjord} proposed a neural ODE-based generative model—the FFJORD—to solve inverse problems; Quaglino et al \cite{quaglino2019snode} used a higher-order approximation of the states in a neural ODE,and proposed the SNet to accelerate computation. Further algorithmic improvements to the framework were presented by YAN et al \cite{yan2019robustness} and Anumasa et al \cite{anumasa2021improving}, exploring the robustness properties of neural ODEs. Effective neural ODE architectures remain the subject of ongoing research — see for example \cite{dupont2019augmented,gholami2019anode,zhuang2020adaptive}.

Training neural ODEs consists of minimizing a loss function over the network weights subject to the nonlinear ODE constraint. To some extent, training can be seen as an optimal control problem. Applying optimal control theory to improve the training has become an appealing research area and more attention has been paid in recent years. For example, Pontryagin’s maximum principle has been used to efficiently train networks with discrete weights \cite{li2017maximum}, multigrid methods have been proposed to parallelize forward propagation during training \cite{gunther2020layer}, and analyzing the convergence on the continuous and discrete level has led to novel architectures \cite{benning2019deep}.  Our goal in this paper is to extend this discussion by optimizing the integral interval of ODEs and perform similar experiments for continuous normalizing flows using neural ODEs from a novel perspective.

In summary, our contributions are as follows:


\begin{itemize}
    \item Firstly, we are the first to propose an improved algorithm based on temporal optimization, which is simple yet effective in significantly boosting the training of neural ODEs. We find that the temporal optimization can attain competitive performance compared to original models but with significantly less training time.
    \item Secondly, we introduce temporal regularization and clipping function, which effectively stablize the training process and do not result in degradation of model performance.
    \item Moreover, we optimize the stopping time $T$ alternately with the parameters $\boldsymbol{\theta}$ of the movement speed $f$, and end up with more compatible $T$ and $\boldsymbol{\theta}$ to cause less number of function evaluation (NFE), and also obtain the decreasing training loss.
\end{itemize}

\section{Preliminaries}
\label{sec:related}

\subsection{Background}
The data distributions encountered in real life are usually complicated, causing the essence behind the data difficult to be explored. One way often be used is to introduce the change of variables formula by $\mathbf{z}=g(\mathbf{x})$, and we have 
\begin{equation}
    \label{eq:tcovf}
    \log p_{\mathbf{x}}(\mathbf{x}) = \log q_{\mathbf{z}}(\mathbf{z})+\log \left | \det \nabla g(\mathbf{x}) \right |,
\end{equation}
where $g: \mathbb{R}^D \xrightarrow{} \mathbb{R}^D $ is bijective, $\nabla g$ is the Jacobian of $g$ and $\det (\cdot)$ is its determinant, $p_{\mathbf{x}}(\mathbf{x})$ and $q_{\mathbf{z}}(\mathbf{z})$ are the distributions of $x$ and $z$, respectively. \textit{}In this way, we can warp the distribution $p_{\mathbf{x}}(\mathbf{x})$ into $q_{\mathbf{z}}(\mathbf{z})$.

In practice, the computational cost to calculate the determinant is $\mathcal{O}(D^3)$, which is the main bottleneck of using Eqn.(\ref{eq:tcovf}). Alteratively, Chen et al \cite{chen2018neural}  use continuous normalizing flow (CNF) to characterize the recursively continuous transformations instead of characterizing $g$ directly in (\ref{eq:tcovf}), and it's computational cost is $\mathcal{O}(D^2)$. For this method, a so called instantaneous change of variables formula is obtained as follows:
\begin{equation}
\label{eq:ticovf}
\begin{gathered}
\partial_{t}\left[\begin{array}{c}
\mathbf{z}(t) \\
\log p (\mathbf{z}(t)) 
\end{array}\right]=\left[\begin{array}{c}
f(\mathbf{z}(t), t; \boldsymbol{\theta}) \\
-\operatorname{Tr}(\mathbf{J}(t, \boldsymbol{\theta}))
\end{array}\right], \\
{\left[\begin{array}{c}
\mathbf{z}(t_0) \\
\log p (\mathbf{z}(t_0)) - \log p_{\mathbf{x}}(\mathbf{x})
\end{array}\right]=\left[\begin{array}{c}
\mathbf{x} \\
\boldsymbol{0}
\end{array}\right], }
\end{gathered}
\end{equation}
where $t \in \left [ t_0, T \right ]$ and $\boldsymbol{\theta}$ are parameters of $f$ which is called the movement speed and is a neural network being trained, $\mathbf{J}(t, \boldsymbol{\theta})=\frac{\partial  f(\mathbf{z}(t), t; \boldsymbol{\theta})}{\partial \mathbf {z}(t)}$ is the partial derivative of $f(\mathbf{z}(t), t; \boldsymbol{\theta})$ w.r.t $\mathbf {z}(t)$. 

Then by integrating across time from $t_0$ to $T$, the change of $\mathbf{z}$ can be derived:
\begin{equation}
\label{infinity}
\begin{split}
    \mathbf{z}(T) 
    &= \mathbf{z}(t_{0}) + \int_{t_{0}}^{T}f(\mathbf{z}(t),t;\mathbf{\theta})\mathrm{d}t \\
    &\approx \mathbf{z}(t_{0}) + \sum_{i=1}^{N} f(\mathbf{z}(t_{i}),t_{i};\mathbf{\theta})\Delta t_{i}, 
\end{split}
\end{equation}
where $N$ denote the number of function evaluations (NFEs).

If training the dynamics (\ref{eq:ticovf}) from the perspective of maximum likelihood estimation, we can switch the estimation of likelihood from $\mathbf{x}$ to $\mathbf{z}$. If $\mathbf{z}$ is an isotropic Gaussian variable, then the likelihood of $\mathbf{x}$ can be computed easily by integrating across time:
\begin{equation}
\label{eq:objective}
\begin{split}
\mathop{\min}\limits_{\boldsymbol{\theta} \in \boldsymbol{\Theta}} L(\boldsymbol{\theta}) 
&= -\mathbb{E}_{p_{\mathbf{x}}}\{\log p(\mathbf{z}(t_{0}); \boldsymbol{\theta})\} \\
&= -\mathbb{E}_{p_{\mathbf{x}}}\left \{\log p(\mathbf{z}(T); \boldsymbol{\theta}) + 
\int_{t_0}^T Tr(\mathbf{J}(t, \boldsymbol{\theta}) )dt\right \},
\end{split}
\end{equation}
where $\mathbf{x}=\mathbf{z}(t_{0})$. 



Furthermore, Grathwohl et al \cite{grathwohl2018ffjord} use the  Hutchinson's trace estimator \cite{hutchinson1989stochastic} and Onken et al \cite{onken2020ot} design a refined network structure to compute the trace in Eqn.(\ref{eq:ticovf}), where the cost are both reduced to $\mathcal{O}(D)$.

Despite the calculation of the log determinant for a single $f$ becomes faster, there still remain some hurdles causing the total evolutionary time unacceptable, like complex structure of $f$ and undesirable large NFEs that increase over time. 

To accelerate the training process of CNF, Finlay et al \cite{finlay2020train} and Onken et al \cite{onken2020ot} introduce several regularization related to $f$ based on the optimal transport (OT) theory. Both of them add transport loss $OT(\boldsymbol{\theta})$ to the objective function, which can be described as:
\begin{equation}
    \label{regularity}
    OT(\boldsymbol{\theta})=\int_{t_0}^{T}\int_{\mathbb{R}^D} \left \| f(\mathbf{z}(t),t;\boldsymbol{\theta}) \right \|^{2}p(\mathbf{z}(t))\mathrm{d}\mathbf{z}\mathrm{d}t.
\end{equation}
Hence, the objective function becomes:
\begin{equation}
\label{eq:objective2}
\mathop{\min}\limits_{\boldsymbol{\theta} \in \boldsymbol{\Theta}} \{L(\boldsymbol{\theta}) + OT(\boldsymbol{\theta})\}.
\end{equation}

Besides, they treat the evolutionary process of $\mathbf{z}$ as the motion of particles and limit the speed of particles $f(\mathbf{z}(t),t;\boldsymbol{\theta})$ at $t$ from different angles.

However, within the above methods, the stopping time, $T$, is treated as a fixed hyperparameter. Given $T$, they have been trying to find the optimal dynamics depending on network weights $\boldsymbol{\theta}$.

\subsection{Related work}

\noindent \textbf{Finite Flows} \quad Normalizing flows \cite{tabak2013family,rezende2015variational,papamakarios2019normalizing,kobyzev2020normalizing} use a finite number of transformations to construct differentiable bijections between complex unknown distributions and simple distributions. NICE \cite{dinh2014nice} and REALNVP \cite{dinh2016density} first use coupling layers to construct the transformation, thus ensuring the reversibility of the model. Improved from REALNVP, a $1 \times 1$ convolution has been introduced in GLOW  \cite{kingma2018glow} to increase the flexibility of the model. Then, an attention mechanism has been developed in FLOW$++$ \cite{ho2019flow++} to obtain a more expressive architecture. An autoregressive structure has been proposed in IAF \cite{kingma2016improved} and MAF \cite{papamakarios2017masked} to enhance the expressiveness of the model. Benefiting from numerous improvements in autoregressive flows  \cite{huang2018neural,durkan2019neural,ziegler2019latent,wehenkel2019unconstrained}, its expressive power is gradually recognized in flow-based models.

~\\
\noindent \textbf{Infinitesimal Flows} \quad Inspired from the presentation of Resnet \cite{he2016deep}, many recent works \cite{salimans2015markov,chen2018neural} have used ordinary differential equations to construct invertible transformations between random variables. A more flexible architecture and lower computational complexity of Jacobian determinant are obtained in FFJORD \cite{grathwohl2018ffjord} by means of unbiased trace estimation. Augmented-NODE \cite{dupont2019augmented} and NANODE \cite{davis2020time} used increased dimensionality to lift the restriction that the trajectories of ordinary differential equations cannot intersect. Most continuous flows back-propogate and update the gradient via the adjoint method, saving memory while back-propogating inaccurate state values. Some models introduce a check-point mechanism to store some of the state nodes during forward propogation to solve the backward integral accurately \cite{gholami2019anode,zhuang2020adaptive,zhuang2021mali}. These infinite-depth approaches from a novel perspective theoretically bridges the gap between deep learning and dynamical systems. In addition, Neural ODEs have shown great promise for numerous tasks such as image registration \cite{xu2021multi}, video generation \cite{park2020vid}, reinforcement learning \cite{du2020model} and system identification \cite{quaglino2019snode}. Recent work has also extended neural ODEs to stochastic differential equations \cite{liu2019neural,oganesyan2020stochasticity}, Riemannian manifold \cite{mathieu2020riemannian}, Bayesian learning frameworks \cite{bhouri2021gaussian} and graph-structured data \cite{huang2021coupled}. 

~\\
\noindent \textbf{Flows with Optimal Transport} \quad To enforce straight trajectories and accelerate training, RNODE \cite{finlay2020train} and OT-FLOW \cite{onken2020ot} regularized the FFJORD model by adding a transport cost in the form of $L_{2}$ norm to the original loss function. RNODE also introduced the Frobenius norm to stablize training. Furthermore, Tay-NODE \cite{kelly2020learning} generalized the form of the $L_{2}$ norm and obtained a regularization term of arbitrary order, but is slower to train than RNODE because of the extra computational cost introduced. TNODE \cite{huang2020accelerating}, on the other hand, proposed a novel view to regularize trajectories into polynomials. STEER \cite{ghosh2020steer}, similar to our method, also optimizes time, but only by random sampling of end time, whereas our method optimizes by coordinate descent as described in Section \ref{3.1}.

\begin{algorithm}
\renewcommand{\algorithmicrequire}{
\textbf{Input:}}
\renewcommand{\algorithmicensure}{
\textbf{Output:}}
\caption{log-density estimaion using the TO-FLOW}
\label{algorithm1}
\begin{algorithmic}
\Require dynamics $f_{\boldsymbol{\theta}}$, start time $t_{0}$, initial stopping time $T_{0}$, minibatch of samples $\mathbf{x}$, number of iterations $n$, network optimizer $\mathcal{P}$, temporal optimizer $\mathcal{Q}$.
\State \textbf{Initialize:} $\mathbf{z}(t_{0}) = \mathbf{x}, T = T_{0}$
\For {$i=1 \to n$}
\State $[\mathbf{z}(T), -\hat{\text{Tr}}] \gets \text{odeint}(f_{\boldsymbol{\theta}}, [\mathbf{x}, \boldsymbol{0}], t_{0}, T)$ $\;$  $\Diamond$ Solve the ODE 
\State $\hat{\boldsymbol{\theta}} \gets \mathcal{P}({\nabla_{\boldsymbol{\theta}} \mathbb{E}_{p_{\mathbf{x}}} \{{-\log p(\mathbf{z}(t_{0}))}}\}, \boldsymbol{\theta})$ $\;$  $\Diamond$ Update the network weights
\State $\hat{T} \gets \mathcal{Q}(\partial_{T} \mathbb{E}_{p_{\mathbf{x}}} \{-\log p(\mathbf{z}(t_0))\}, T)$ $\;$  $\Diamond$ Update the stopping time
\EndFor
\Ensure final dynamics $f_{\hat{\boldsymbol{\theta}}}$, stopping time $\hat{T}$
\end{algorithmic}
\end{algorithm}

\begin{figure*}
\begin{center}

\includegraphics[width=0.99\textwidth]{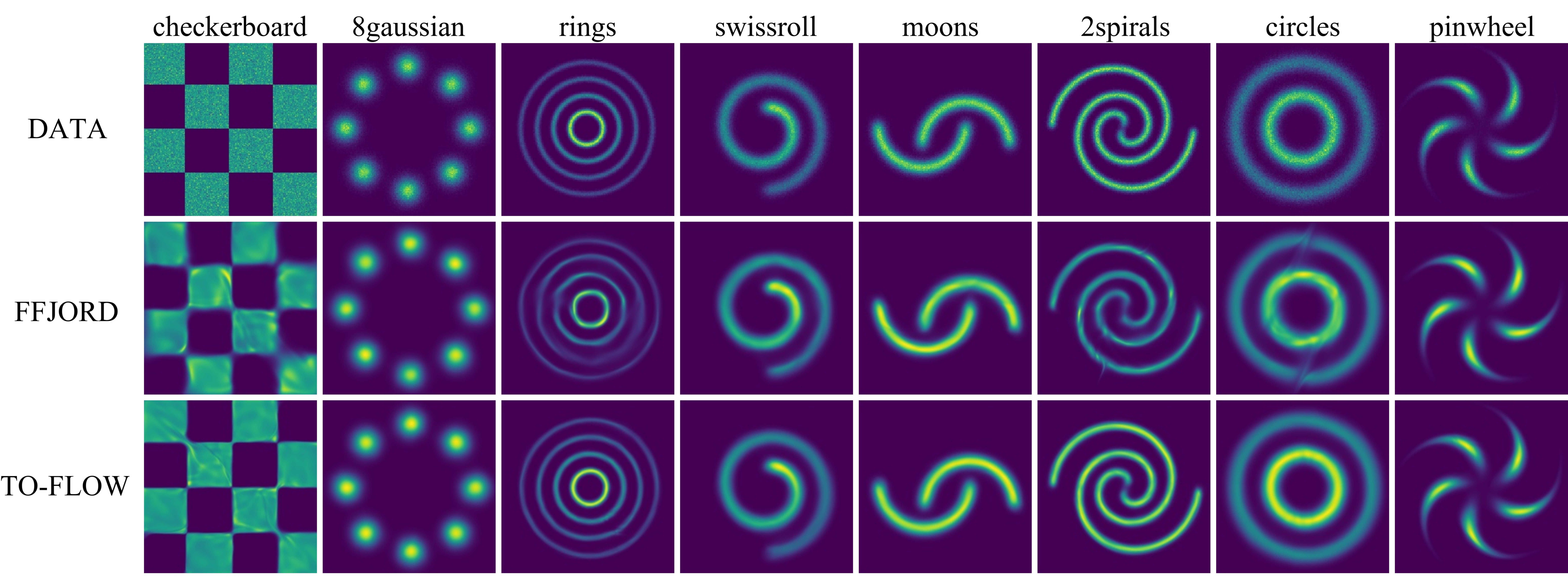}
\hfill
\end{center}
\caption{Comparison of FFJORD and TO-FLOW on 2-dimensional distributions.}
\label{fig:toy-result}
\end{figure*}
\begin{figure*}
\begin{center}

\includegraphics[width=0.67\textwidth]{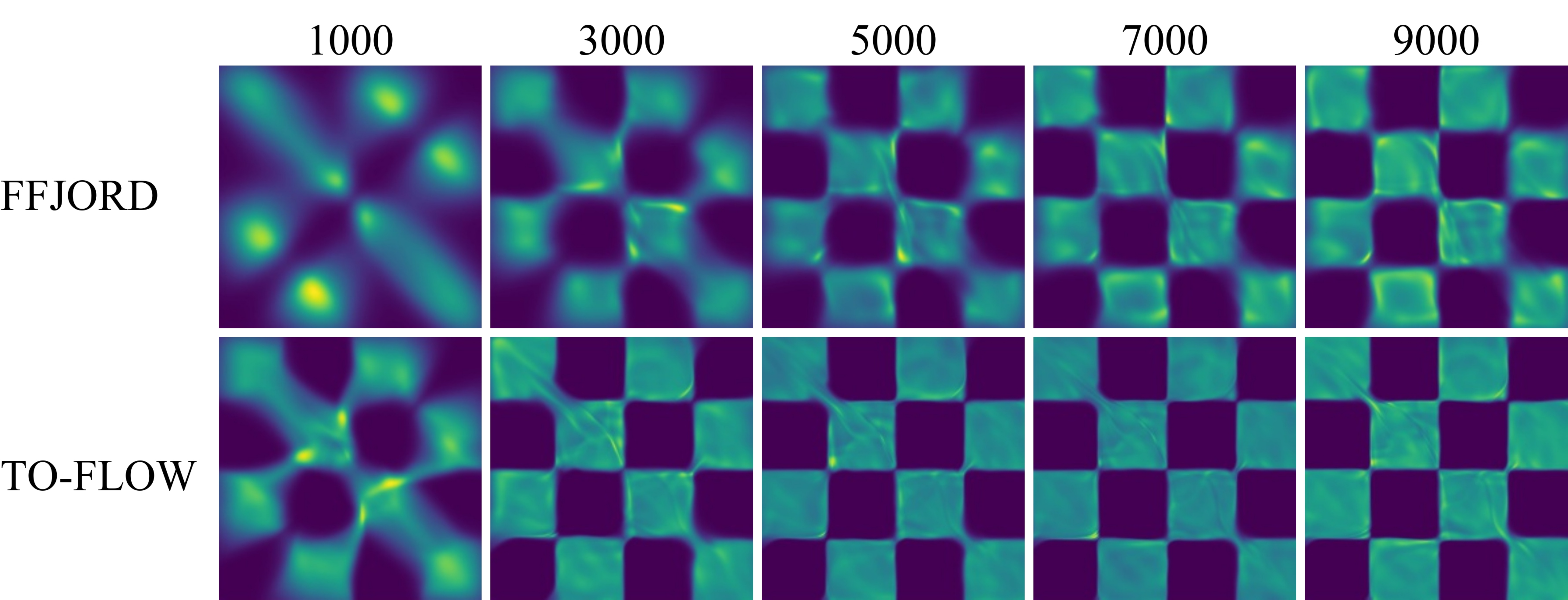}
\hfill
\end{center}
\caption{Comparison of FFJORD and TO-FLOW on checkerboard data set. The numbers at the top of the images represent the number of iterations of the model.}
\label{fig:stage-checkerboard}
\end{figure*}
\section{Method}
\label{sec:method}
Motivated by the similarities between training CNFs and solving OT problems \cite{benamou2000computational,peyre2019computational}, some previous works regularize the minimization problem (\ref{regularity}) to enforce a straight trajectory and significantly faster training \cite{finlay2020train,huang2020accelerating,kelly2020learning}.

From another perspective, if we express Eqn. (\ref{eq:ticovf}) explicitly by the first-order Euler method: 
\begin{equation}
\label{euler}
    \mathbf{z}(t_{i}+\Delta t_{i}) = \mathbf{z}(t_{i}) + f(\mathbf{z}(t_{i}),t_{i};\boldsymbol{\theta}) \Delta t_{i} 
\end{equation}

\begin{equation}
\label{add}
    T = t_{0} + \sum_{i=1}^{n} \Delta t_{i},
\end{equation}

\noindent where $n$ denote total number of time steps. It is clear that total evolutinary time $T-t_{0}$ interacts with $f(\mathbf{z}(t),t;\boldsymbol{\theta})$ to infuence the evolution of $\mathbf{z}(t)$ in an intricate way. Without placing demands on the total evolutionary time $T-t_{0}$, an under-regularized trajectory is formed and results in unnecessarily large training time \cite{ghosh2020steer}.

How can the regularization of the total evolutionary time be designed? The formulation of continuous normalizing flows is given by (\ref{infinity})
where $t_{0}$ and $T$ are both hyperparameters that fixed before training. One intuitive approach is to optimize $t_{0}$ and $T$ together to find an appropriate integral horizon. For simplicity, we fix $t_{0}$ and only optimize $T$ (for the derivation of the generic form, see App.A). 

Then combing with the optimization of the moving speed depending on $\theta$, we could revise the optimization problem \ref{eq:objective} as follows:
\begin{equation}
\label{eq:objective3}
\mathop{\min}\limits_{\boldsymbol{\theta} \in \boldsymbol{\Theta}, T \in \mathbb{R}} \{L(\boldsymbol{\theta}, T)\}.
\end{equation}


\subsection{Coordinate descent}
\label{3.1}
The problem is split into two smaller subproblems: one trains the network weights $\boldsymbol{\theta}$, the other optimizes the stopping time $T$ to form an appropriate trajectory. In the following, the details of each subproblem are discussed.

~\\
\noindent {\bfseries{Step 1: Training of the network weights.}}  The initial network weights are chosen randomly, then updated by fixing the stopping time $\Tilde{T}$ and solving the subproblem:
\begin{equation}
    \label{minimizetheta}
    \mathop{\min}_{\boldsymbol{\theta} \in \boldsymbol{\Theta}} L(\boldsymbol{\theta},\Tilde{T}) =  -\mathbb{E}_{p_{\mathbf{x}}}\{\log p(\mathbf{z}(t_0); \boldsymbol{\theta})\},
\end{equation}
Once the objective function is solved, then we calculate the gradient ${\nabla_{\boldsymbol{\theta}} L(\boldsymbol{\theta},\Tilde{T})}$ to update the network weights $\boldsymbol{\theta}$ with some general SGD-like optimizer $\mathcal{P}$, such as Adam \cite{kingma2014adam}.

~\\
\noindent {\bfseries{Step2: Coordinate descent on evolutionary time.}}  Once the network weights $\boldsymbol{\theta}$ are updated, we then fix the updated network weights $\Tilde{\theta}$, and solve the subproblem:
\begin{equation}
\begin{split}
    \label{minimizeT}
    \mathop{\min}_{T \in \mathbb{R}} L(\boldsymbol{\Tilde{\theta}},T) &=  -\mathbb{E}_{p_{\mathbf{x}}}\{\log p(\mathbf{z}(t_0); \boldsymbol{\Tilde{\theta}})\}  \\
        &= -\mathbb{E}_{p_{\mathbf{x}}}\left \{\log p(\mathbf{z}(T); \boldsymbol{\Tilde{\theta}}) + 
\int_{t_0}^T Tr(\mathbf{J}(t, \boldsymbol{\Tilde{\theta}}) )dt\right \},
\end{split}
\end{equation}
The stopping time $T$ is updated by calculating the derivative of $L(\boldsymbol{\Tilde{\theta}},T)$ with respect to $T$:
\begin{equation}
\begin{split}
    \label{derivative}
    \frac{\partial L(\boldsymbol{\Tilde{\theta}},T)}{\partial T} &= -\frac{\partial \mathbb{E}_{p_{\mathbf{x}}}\{\log p(\mathbf{z}(T);\boldsymbol{\Tilde{\theta}})\}}{\partial T} 
    - \mathbb{E}_{p_{\mathbf{x}}}\{ Tr(\mathbf{J}(T, \boldsymbol{\Tilde{\theta}}))\} \\
\end{split}
\end{equation}

\begin{table*}[t]
\centering
\resizebox{\linewidth}{!}{ 
\centering
\begin{tabular}{llcccccc}   
\toprule
Data Set & Model & Bits/dim & Param & Time(h) & Iter & Time/Iter(s) & NFE\\
\midrule
 & FFJORD & 1.017 & 400K & 79.641 & 60K & 6.409 & 750.67 \\
MNIST & STEER & 1.024 & 400K & 138.212 & 60K & 12.368 & 1265.48 \\
 & TO-FLOW (ours) & 1.026 & 400K & \textbf{46.363} & 60K & \textbf{3.353} & \textbf{396.81} \\
\hline

 & FFJORD & 2.806 & 400K & 87.845 & 60K & 7.010 & 811.40 \\
Fashion-MNIST & STEER & 2.803 & 400K & 147.197 & 60K & 12.405 & 1308.82 \\
 & TO-FLOW (ours) & 2.807 & 400K & \textbf{63.482} & 60K & \textbf{5.415} & \textbf{513.79} \\

\hline

 & FFJORD & 3.414 & 670K & 108.314 & 50K & 10.299 & 1228.04 \\
CIFAR-10 & STEER & 3.424 & 670K & 168.649 & 50K & 15.502 & 1749.17 \\
 & TO-FLOW (ours) & 3.429 & 670K & \textbf{82.607} & 50K & \textbf{7.373} & \textbf{716.85} \\

\bottomrule
\end{tabular}
             } 
\caption{Density estimation on image data sets. We present the testing loss (Bits/dim), number of parameters (Param), total training time (Time), total number of iterations (Iter), average time per iteration (Time/Iter) and average number of function evaluations (NFE). We use moving average instead of summation average \cite{grathwohl2018ffjord}.}
\label{sota}
\end{table*}


However, partial derivative $\frac{\partial \log p}{\partial T}$ cannot be derived directly, since we only encode the stopping time $T$ onto the channels of the feature map without introducing a direct correspondence. Instead, we introduce the chain rule to obtain a feasible calculation:
\begin{equation}
    \label{chain}
        \frac{{\partial \mathbb{E}_{p_{\mathbf{x}}}\{\log p(\mathbf{z}(T);\boldsymbol{\Tilde{\theta}})}\}}{\partial T} =  \mathbb{E}_{p_{\mathbf{x}}} \left \{ \frac{ \partial  \log p(\mathbf{z}(T);\boldsymbol{\Tilde{\theta}})}{\partial \mathbf{z}(T)} \circ \frac{\partial \mathbf{z}(T)}{\partial T} \right \},
\end{equation}
where $\circ$ denotes the dot product and $\frac{\partial \mathbf{z}(T)}{\partial T}=f(\mathbf{z}(T),T;\boldsymbol{\Tilde{\theta}})$, which is relatively well calculated.

Once the derivative is calculated, we update $T$ also with some SGD-like optimizer $\mathcal{Q}$, such as Adam \cite{kingma2014adam}:
\begin{equation}
    \label{update}
    \hat{T} = \mathcal{Q} \left (\frac{\partial L(\boldsymbol{\Tilde{\theta}},T)}{\partial T},T \right ),
\end{equation}
where $\widetilde{T}$ denote the stopping time updated in one iteration. Pseudo-code of our method is given in Algorithm \ref{algorithm1}.

\begin{figure}[t]
\begin{center}

\includegraphics[width=\linewidth]{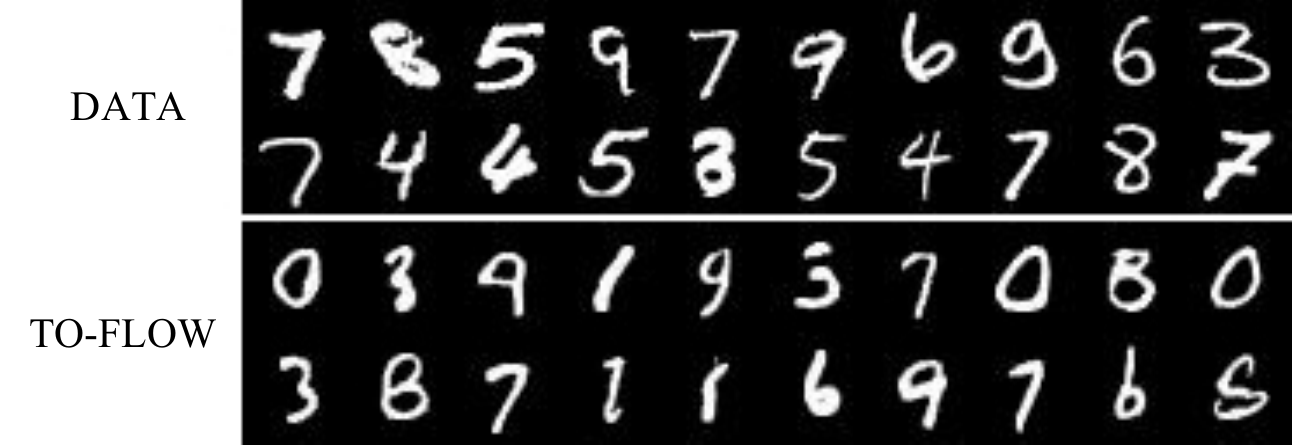}

\end{center}
\caption{Samples of MNIST data set.}
\label{fig:mnist-regularize}
\end{figure}
\begin{figure}[t]
\begin{center}

\includegraphics[width=\linewidth]{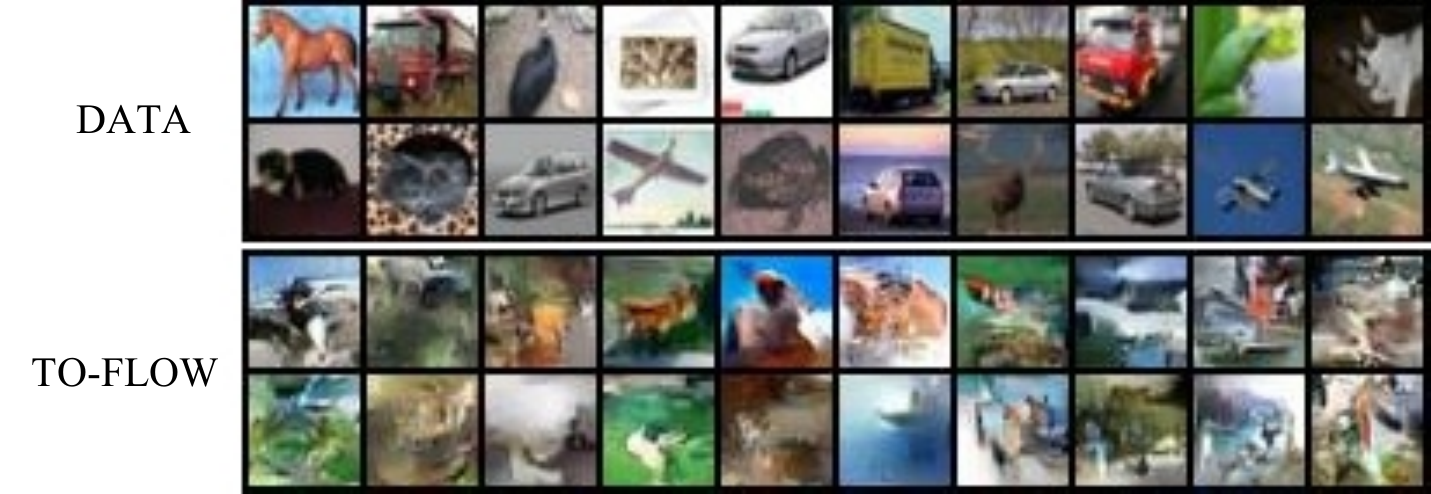}

\end{center}
\caption{Samples of CIFAR-10 data set.}
\label{fig:cifar-regularize}
\end{figure}
\begin{figure}[t]
\begin{center}

\includegraphics[width=\linewidth]{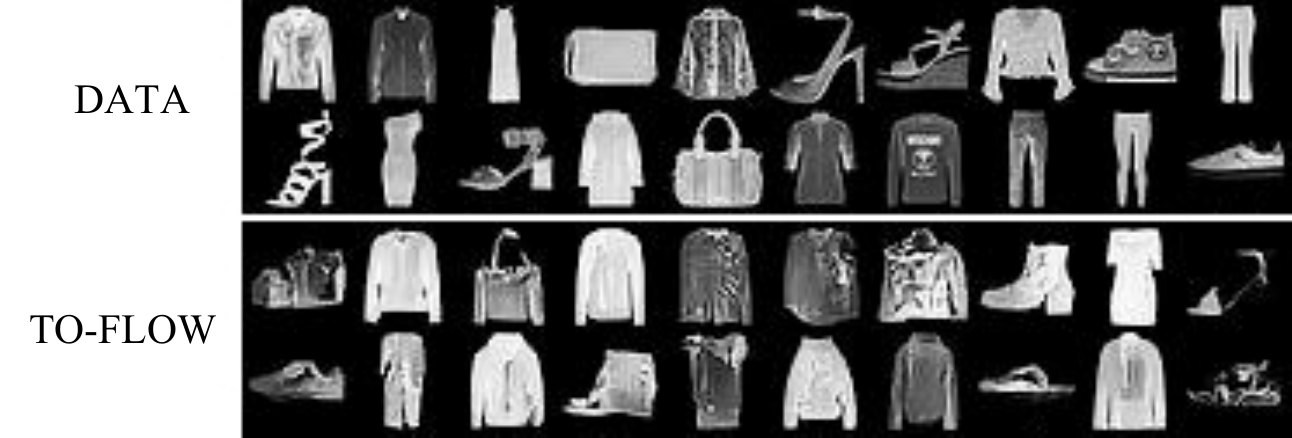}

\end{center}
\caption{Samples of FASHION-MNIST data set.}
\label{fig:fashion-regularize}
\end{figure}
\subsection{Temporal regularization}
Within our experiments, we find that $T$ changes quickly at the beginning of training. Finlay et al \cite{finlay2020train} and Onken et al \cite{onken2020ot} constrain the movement speed $f(\mathbf{z}(t),t;\boldsymbol{\theta})$ of particles to reduce the loss of transmission from the perspective of OT theory. Inspired by their work, we add constraints to the total evolutionary time $T-t_{0}$ to stablize training. This trick is called temporal regularization (TR), which can be described as:
\begin{equation}
    \label{eq:TR}
    TR(T) = \alpha \cdot |T|,
\end{equation}
where $\alpha$ denotes the power of TR on the training process of CNF. Then the total objective function becomes: 
\begin{equation}
\label{eq:minTR}
\mathop{\min}\limits_{\boldsymbol{\theta} \in \boldsymbol{\Theta}, T \in \mathbb{R}} \{L(\boldsymbol{\theta}, T) + TR(T)\}.
\end{equation}

\begin{figure*}
\begin{center}
\includegraphics[width=.67\columnwidth]{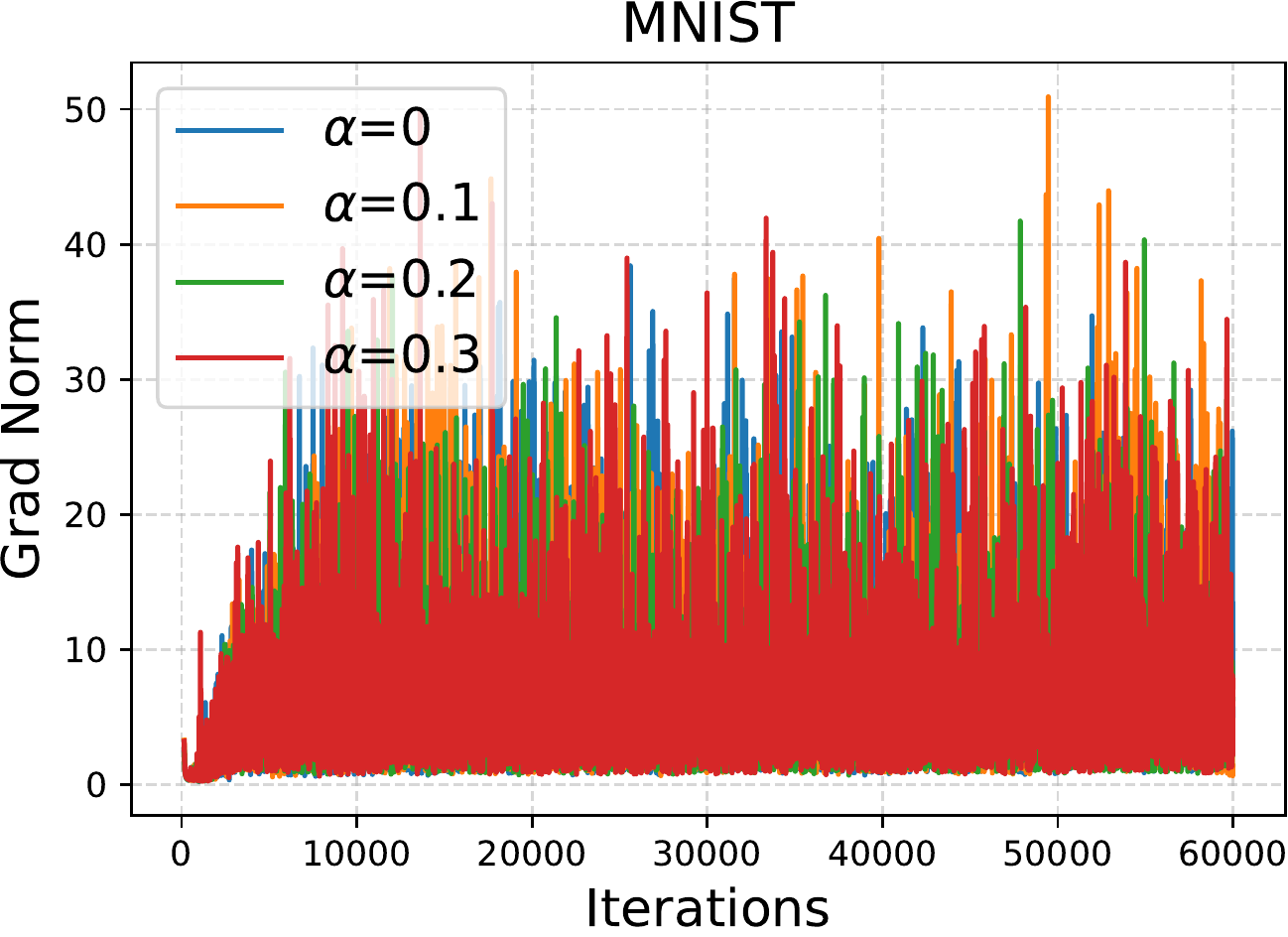}
\hfill
\includegraphics[width=.67\columnwidth]{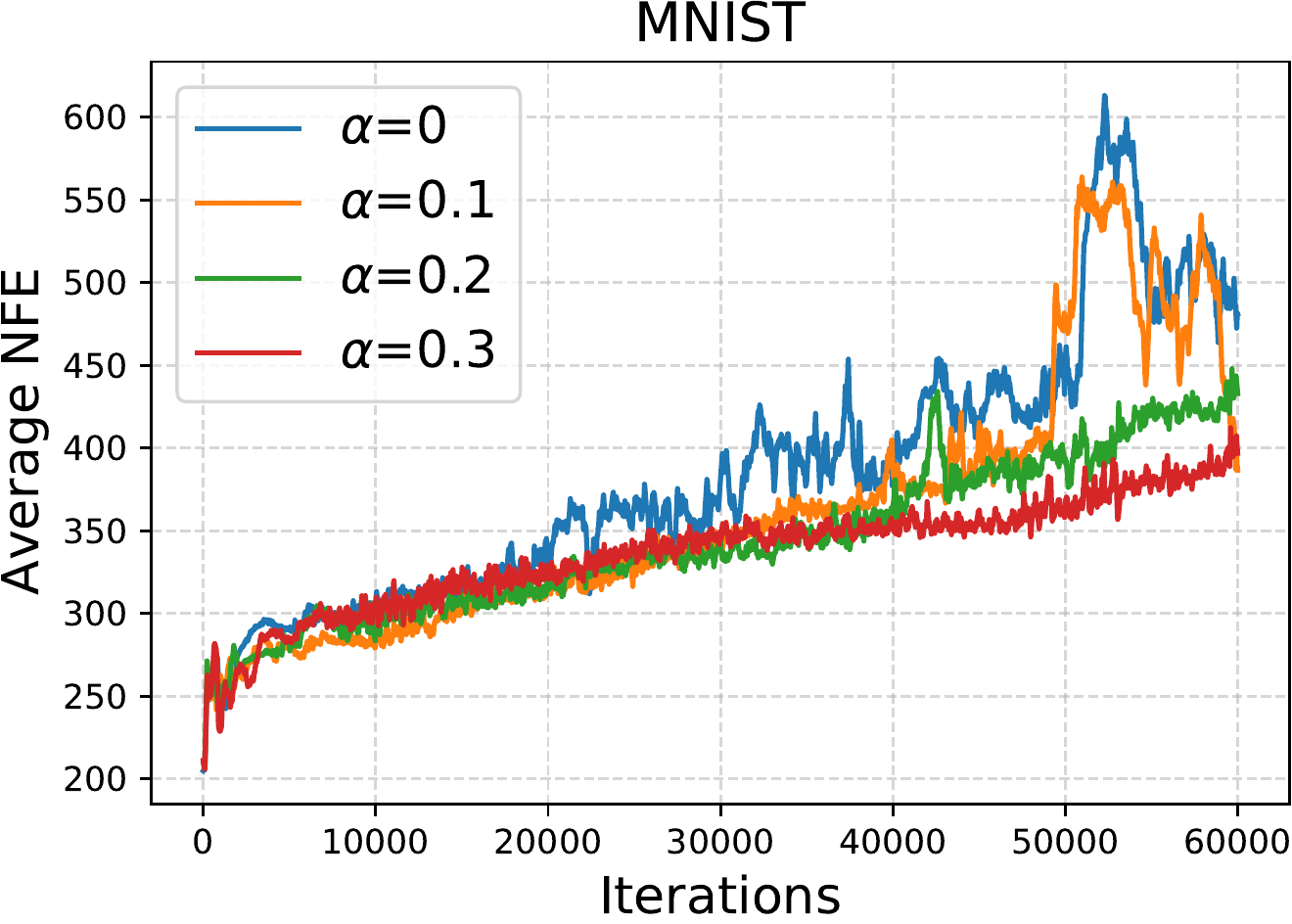}
\hfill
\includegraphics[width=.67\columnwidth]{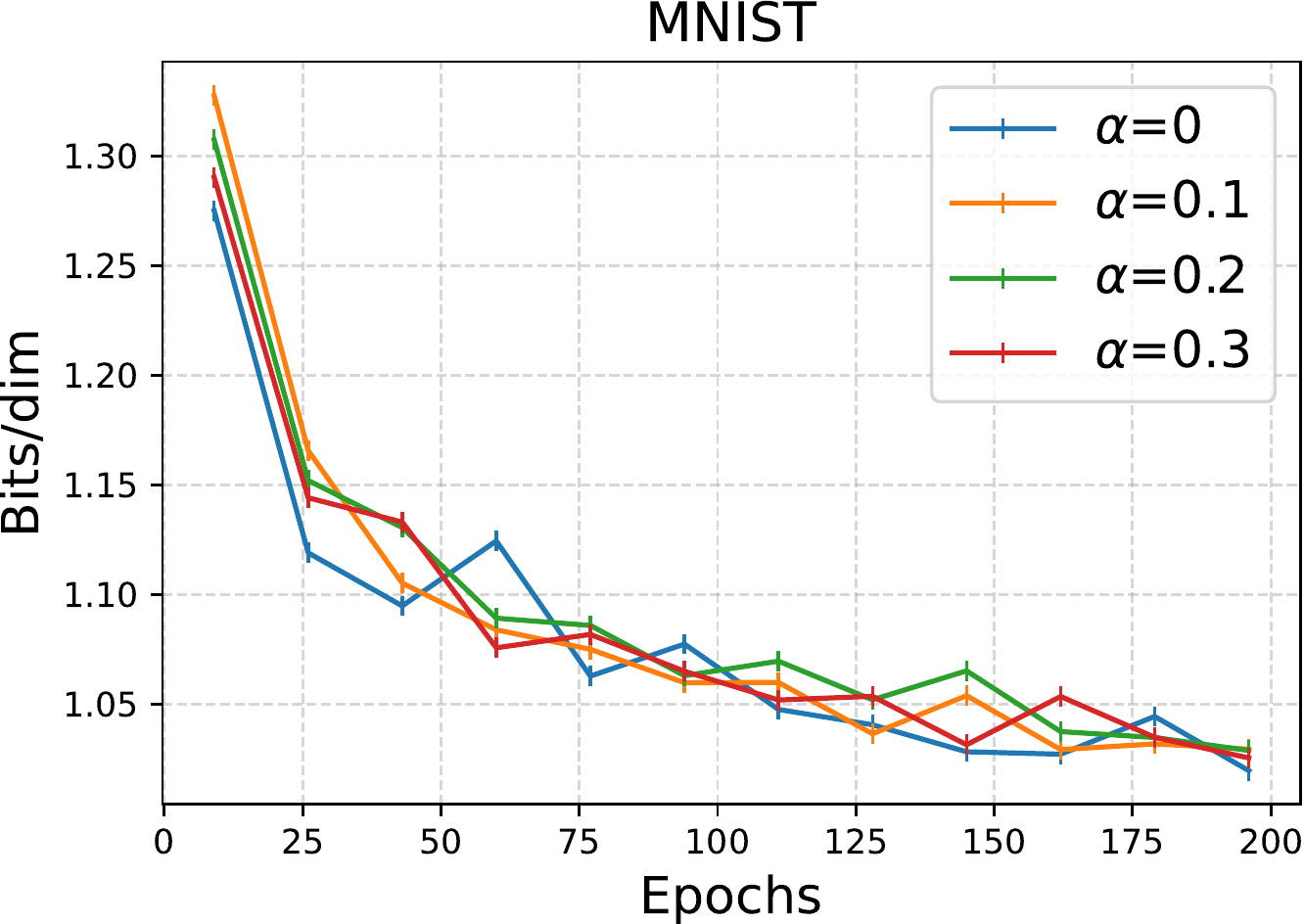}
\end{center}
\caption{Model performance under different temporal regularization on MNIST data set.}
\label{fig:setting-mnist}
\end{figure*}

\begin{figure*}
\begin{center}
\includegraphics[width=.67\columnwidth]{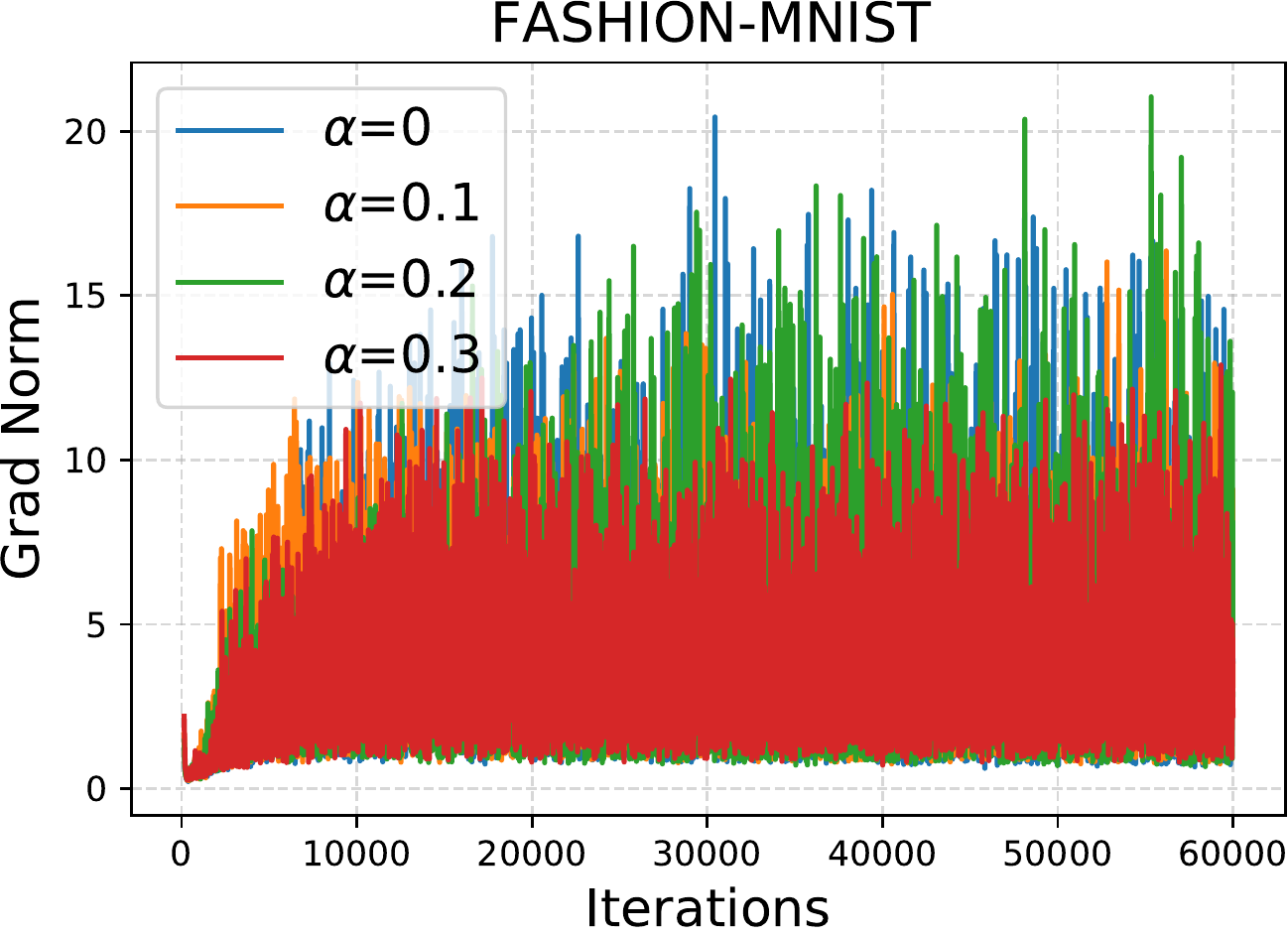}
\hfill
\includegraphics[width=.67\columnwidth]{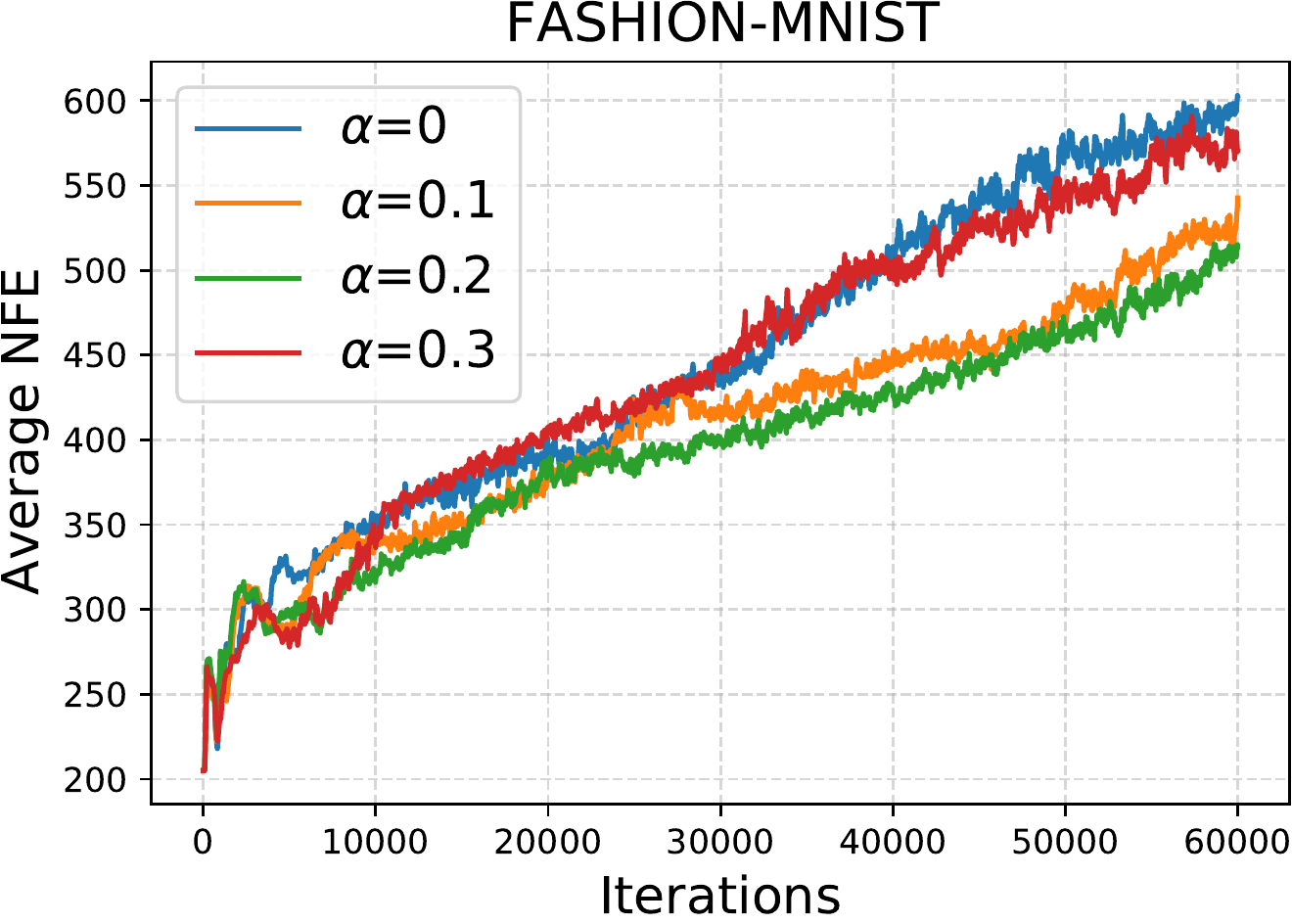}
\hfill
\includegraphics[width=.67\columnwidth]{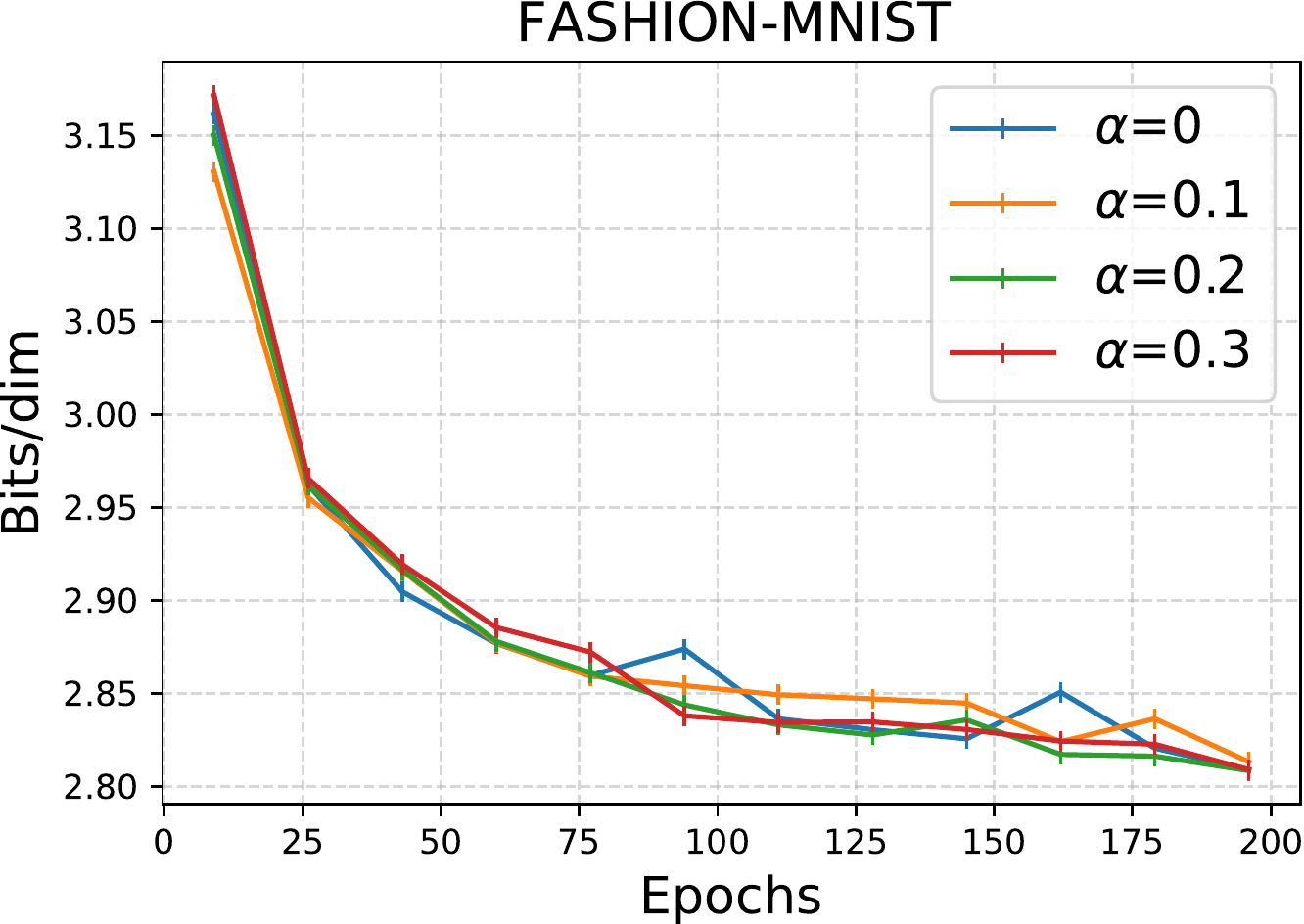}
\end{center}
\caption{Model performance under different temporal regularization on FASHION-MNIST data set.}
\label{fig:setting-fashion}
\end{figure*}

How does TR affect the training process of CNF? It can be seen in the left side of Figure \ref{fig:setting-mnist} and \ref{fig:setting-fashion} that a smaller $\alpha$ can lead to instability in the training process. Hence the hyperparameter $\alpha$ can be measured as the strength of temporal regularization.

We also propose an operation of applying a clipping function to the stopping time $T$. In this case, the $T$ obtained at each iteration of the model is in the interval $[t_0+\varepsilon, 2T_0-t_0-\varepsilon]$, the center of which is $T_0$. Empirically, an obvious advantage of this is that $T$ will not evolve drastically during each iteration. The so called clipping function at each iteration is defined as:
\begin{equation}
\operatorname{Clip}\left( T \right)= \begin{cases}2T_{0} - t_{0} - \varepsilon & T \geq 2T_{0} - t_{0} - \varepsilon \\ t_{0} + \varepsilon & T \leq t_{0} + \varepsilon \\ 
T & \text { otherwise }\end{cases}
\end{equation}
where $\varepsilon$ is the clipping parameter. $t_0$ and $T_0$ denote the initial value of the lower and upper bound of the integral, respectively.


\begin{figure*}
\begin{center}
\includegraphics[width=.67\columnwidth]{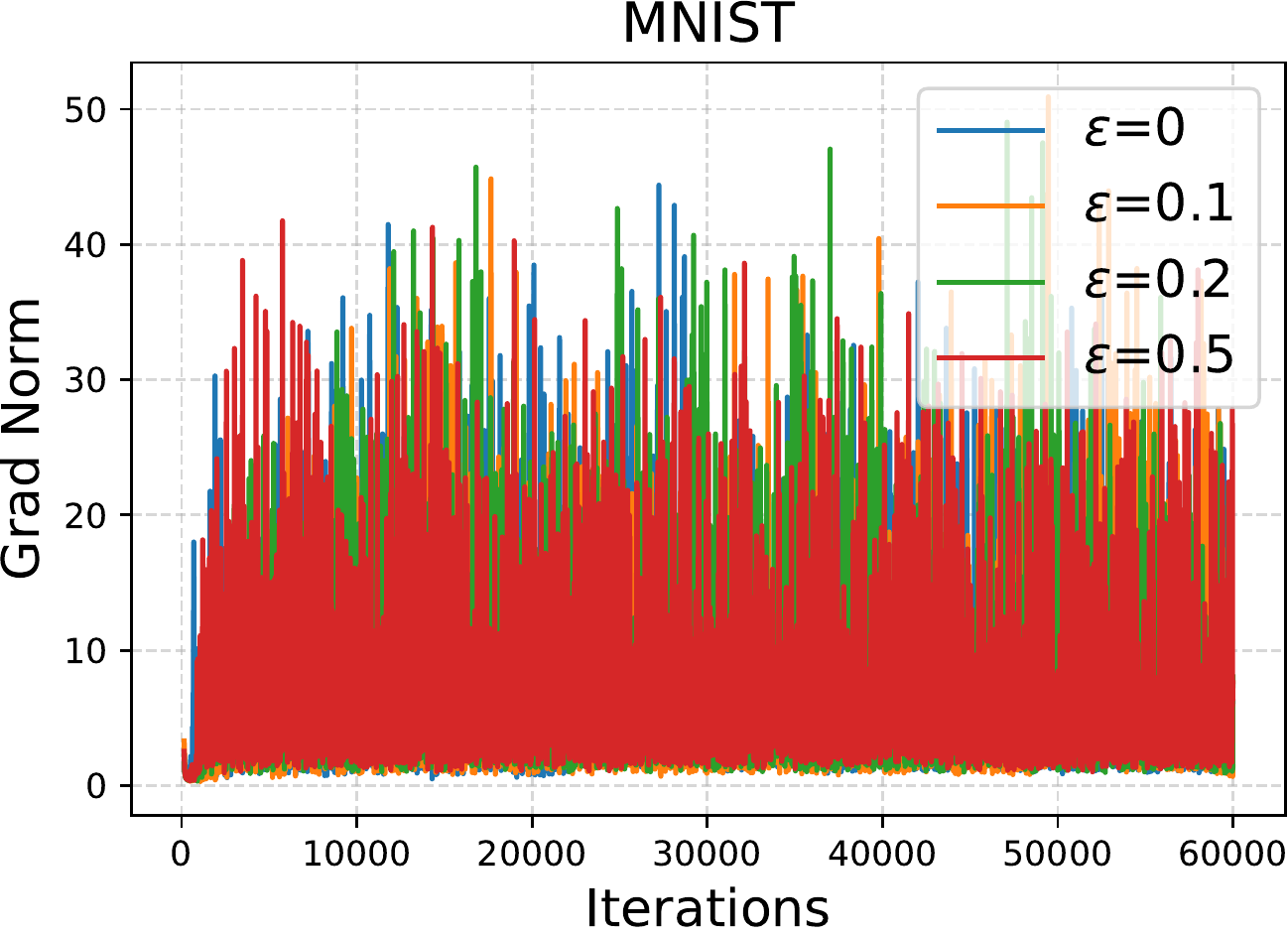}
\hfill
\includegraphics[width=.67\columnwidth]{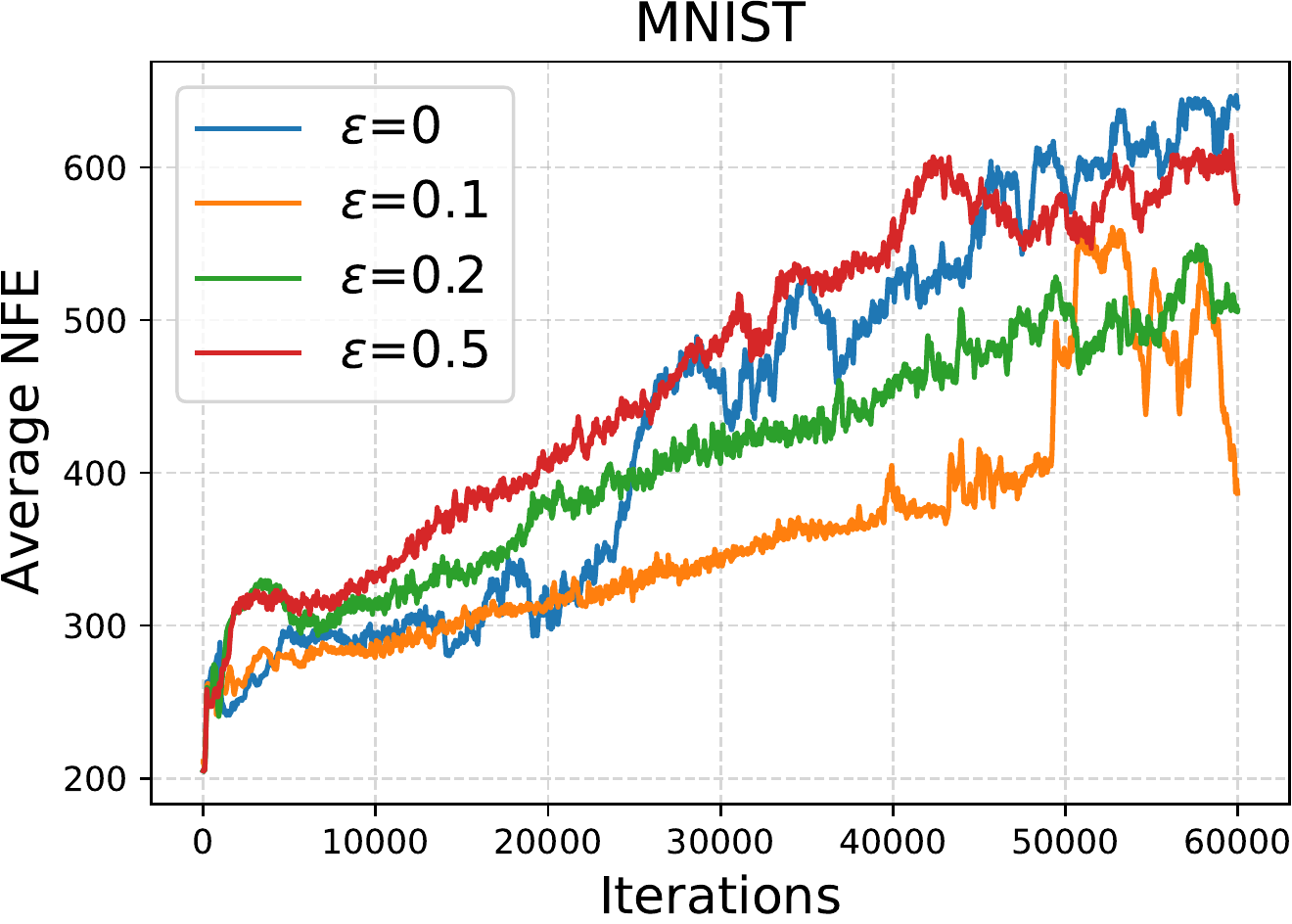}
\hfill
\includegraphics[width=.67\columnwidth]{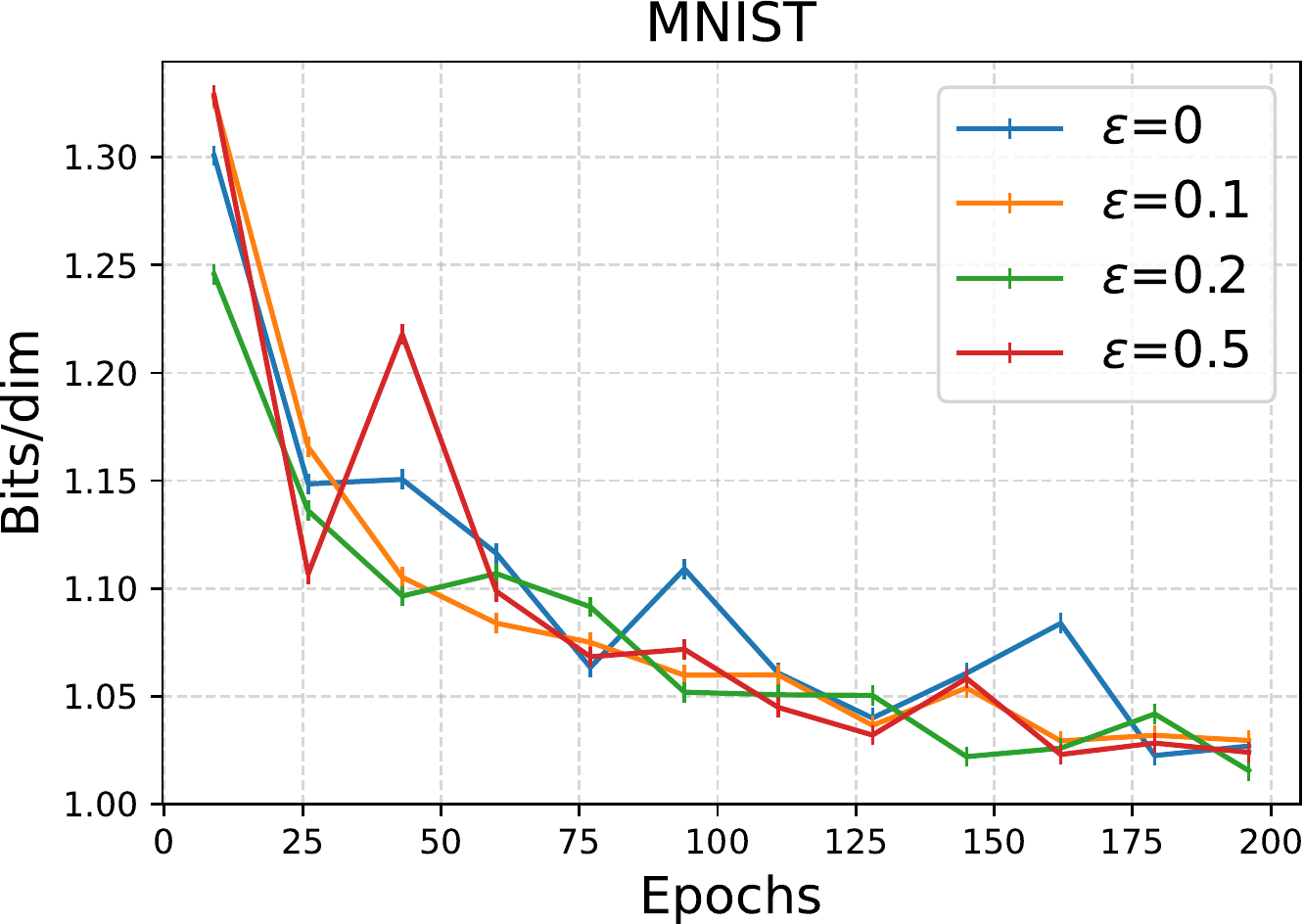}
\end{center}
\caption{Model performance via different clipping function on MNIST data set.}
\label{fig:clip-mnist}
\end{figure*}
\begin{figure*}
\begin{center}
\includegraphics[width=.67\columnwidth]{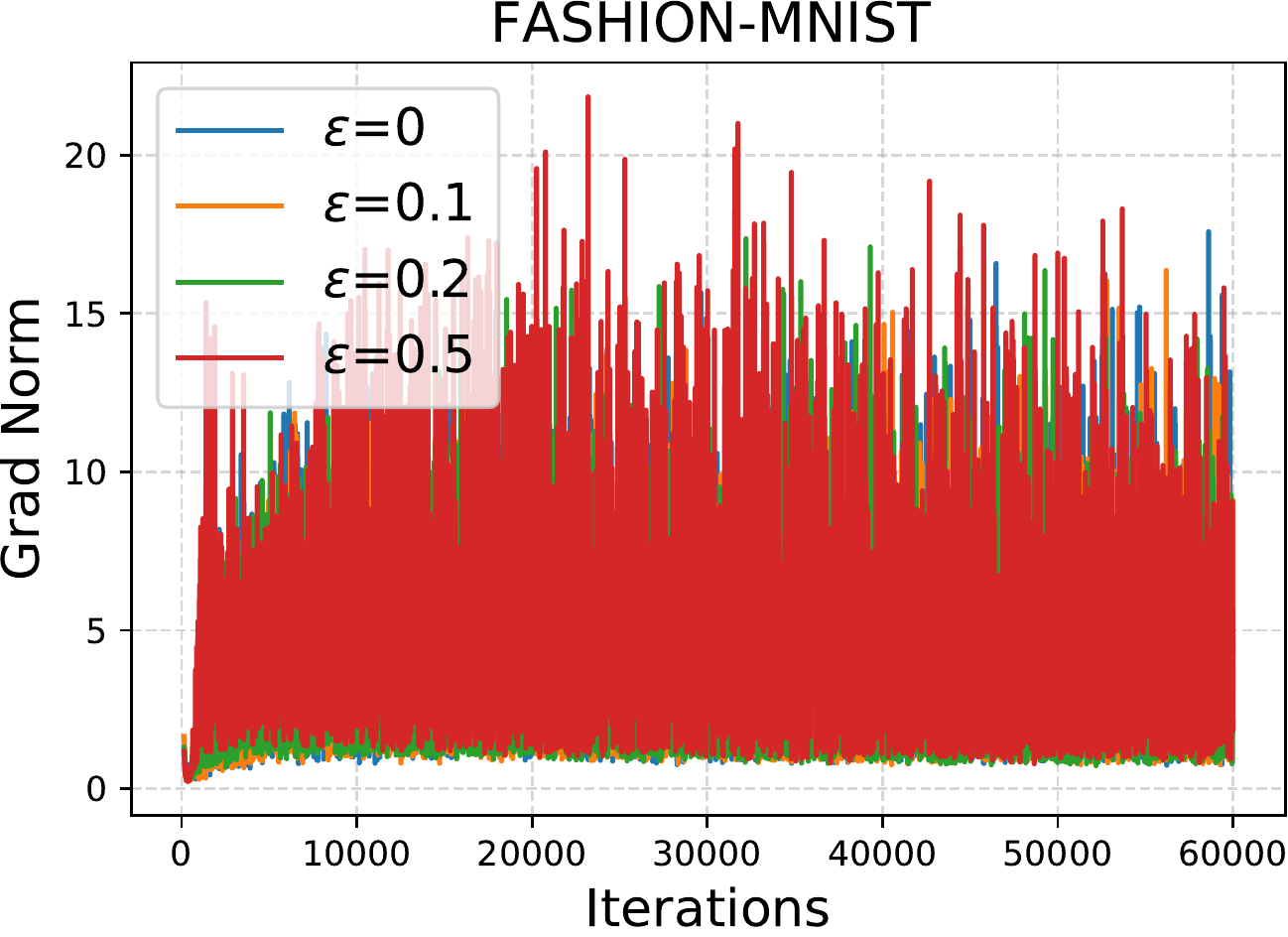}
\hfill
\includegraphics[width=.67\columnwidth]{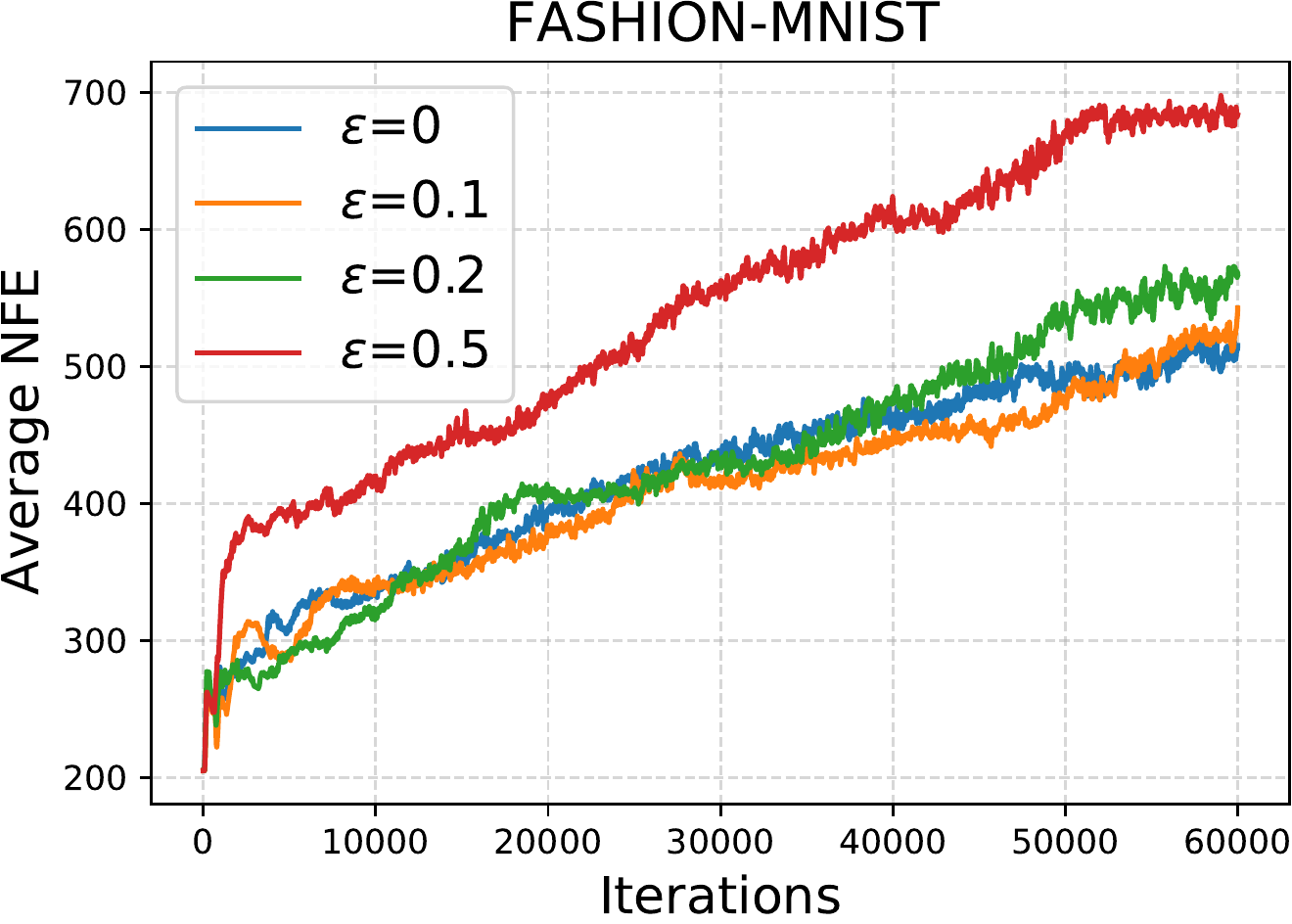}
\hfill
\includegraphics[width=.67\columnwidth]{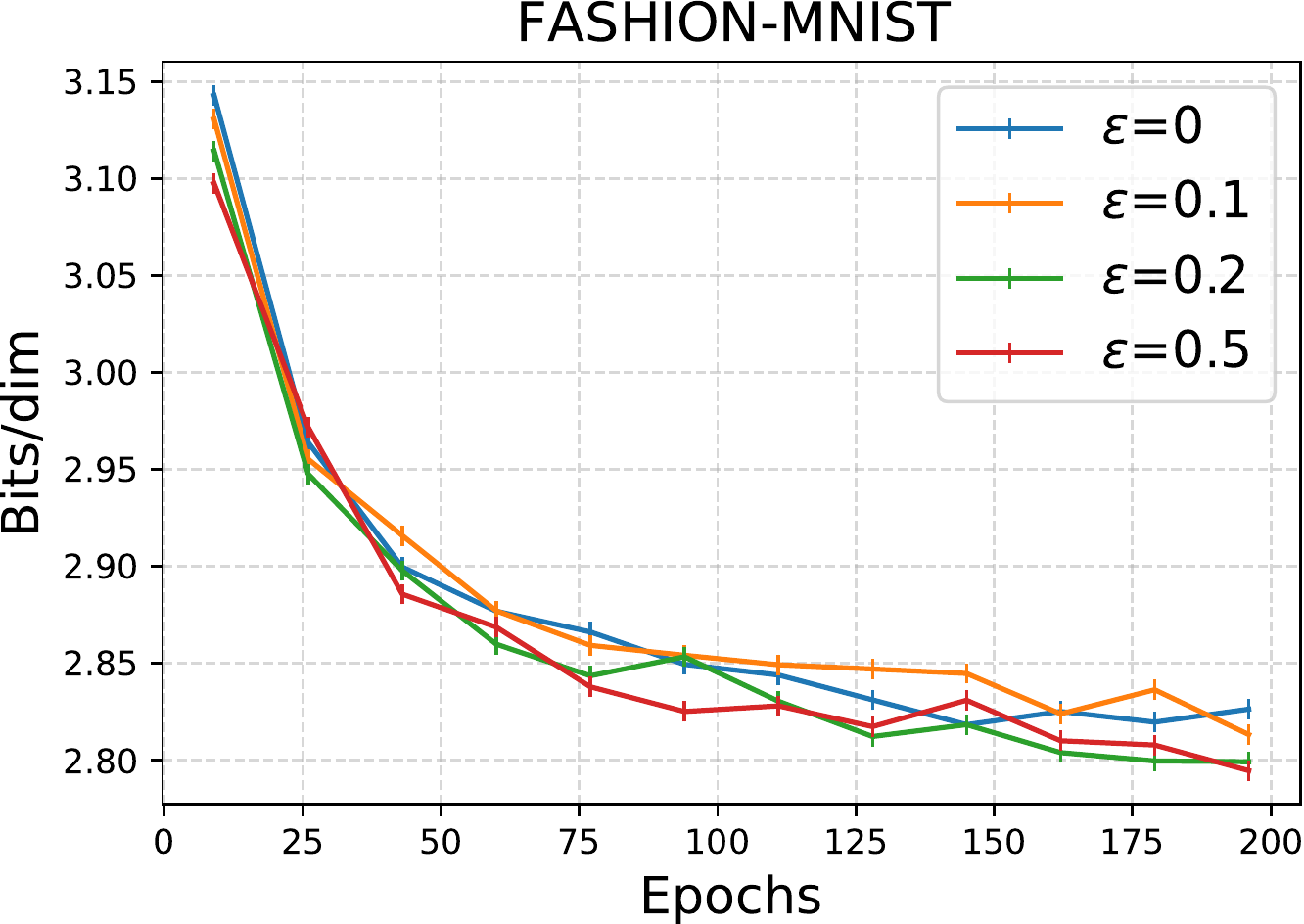}
\end{center}
\caption{Model performance via different clipping function on FASHION-MNIST data set.}
\label{fig:clip-fashion}
\end{figure*}
\section{Experiments}
We demonstrate the benefits of the proposed method on a variety of density estimation tasks. We compare our results with FFJORD \cite{grathwohl2018ffjord}, the baseline of our method, and STEER \cite{ghosh2020steer}, another model that only by random sampling the stopping time.

Two metrics are evaluated, testing loss and training time. We want to see whether our model leads to faster training process compared to FFJORD and STEER in training, while keeping comparable training quality. 

To compare the training speed, we count total training time and average time per training iteration. We also count the average NFE per training iteration. NFE is defined as the number of function of evaluating the right-hand-side of the ODE \ref{eq:ticovf} when solving it. The lower NFE, the faster training speed.

To evaluate the training quality, we compute bits/dim as a metric:
\begin{equation}
    \label{bitsperdim}
    \text{BPD} = -\mathbb{E}_{p_{\boldsymbol{x}}} \left \{\frac{\log \hat{p}(x)/d - \log 256}{\log 2} \right \},
\end{equation}
where $\log \hat{ p}(x) $ denote estimated log-likelihood of our model and $d$ denote the dimension of data. It is a classical metric to measure the approximation of the distribution transformed by flow-based model to the isotropic Gaussian distribution. A low $\text{BPD}$ value means that the model can effectively transform an unknown data distribution into a simple known distribution.

In all experiments, we use exactly the same architectures of neural network as in FFJORD. What we do is integrating our temporal optimization to training process. The experiment settings of temporal optimization are described below. For temporal optimizer, we choose Adam \cite{kingma2014adam} as an optimizer and set the learning rate $lr = 10^{-2}$. The initial value of the stopping time $T_{0}$ is set to be the same fixed value as in FFJORD, which is $0.5$ in toy data and $1$ in image data. The hyperparameter of the temporal regularization is $\alpha = 0.1$. The hyperparameter of the clipping function is $\epsilon = 0.1$. These hyper-parameters above are shared in all experiments. Furthermore, we discuss the influences of different choice of hyperparameters in Section \ref{discussion}

\subsection{Density estimation on toy 2D data}
We first train TO-FLOW on eight simple 2D toy data that serve as standard benchmarks \cite{grathwohl2018ffjord}. In Figure \ref{fig:toy-result}, we show that TO-FLOW can fit both multi-modal and  discontinuous distributions compared against FFJORD by warping a simple isotropic Gaussian.

The distributions of the eight 2D data used for the experiment are shown in the first row of Figure \ref{fig:toy-result}. The learned distributions using FFJORD and our method are shown in the second row and the last row, respectively. Both FFJORD and TO-FLOW trained 10000 iterations for a fair comparison. For checkerboard, rings, 2spirals and circles, our model produces images with a higher degree of reduction, which also shows that our approach can learn multi-modal and discontinuous distributions more efficiently. We compare different stages of FFJORD and TO-FLOW on checkerboard data set in Figure \ref{fig:stage-checkerboard}. For a more detailed comparison of different stages during training, see App.B.

\subsection{Density estimation on image data}
We compare our model's performance on three image data sets: MNIST \cite{lecun1998mnist}, CIFAR-10 \cite{krizhevsky2009learning} and FASHION-MNIST \cite{xiao2017fashion}. On three data sets, we train with a batch size of 200 and train for 200 epochs on a single GPU\footnote{We use Tesla-V100 for all experiments.}.

A comparison of TO-FLOW against FFJORD and STEER\footnote{We obtained results that differ from the original paper. Since the author did not release their code, We will continue to try other strategies to replicate their results.} is presented in Table \ref{sota}. For training speed, our model significantly outperforms FFJORD and STEER. On all data sets, our model uses fewer NFE and a shorter training time, which also leads to a faster convergence of the model. The reduction of total training time ranges from $23.7\%$ to $41.7\%$. On some large data sets, such as CIFAR-10, our model is $23.7\%$ faster than the baseline model, which demonstrates the great potential of our model to scale to larger datasets.

For testing loss (\ref{bitsperdim}), our model is also comparable with FFJORD and STEER, which shows that there is no performance penalty for adding temporal optimization. We also visualize the images generated by our model in Figure \ref{fig:mnist-regularize}, \ref{fig:cifar-regularize} and \ref{fig:fashion-regularize}. 
As can be seen, the introduction of temporal optimization still maintains the quality of image generation. More images generated by different settings can be seen in App.C.

In general, the proposed approach results in comparable performance in tesing loss, but significantly speeds up training. It allows us to use larger network structures and batch sizes, which also preserves the possibility of further performance improvements.
\section{Analysis and discussion}
\label{discussion}
We perform a series of ablation experiments to gain more insight into our model.

\subsection{Stable training via temporal regularization}
Grad norm denote the clipped portion of the norm of the gradient of the network parameters and is often used in training to characterize the stability of the training process. We compare the performance of our model under different coeffieient of temporal regularization on MNIST and FASHION-MNIST. We plot the grad norm in the left side of Figure \ref{fig:setting-mnist} and \ref{fig:setting-fashion}. We plot average NFE and testing loss in the middle and right side of Figure \ref{fig:setting-mnist} and \ref{fig:setting-fashion} respectively to measure the effect of temporal regularization on training speed and density estimation. We find that the introduction of temporal regularization  significantly stablizes the training process while maintaining training speed and the accuracy of density estimation.

\subsection{The choice of clipping function}
The performance of Temporal Optimization is closely related to the choice of clipping parameter $\epsilon$ since it represents the boundary of the evolutionary time $T-t_{0}$. We compare the model performance of different size of clipping parameter $\epsilon$ and plot  in Figure \ref{fig:clip-mnist} and \ref{fig:clip-fashion} respectively. We can conclude that a more compact boundary also leads to more stable training and does not result in degradation of model performance.

\subsection{Future work}
Finlay et al \cite{finlay2020train} and Onken et al \cite{onken2020ot} are dedicated to optimising trajectory spatially, thus constraining it to be straight to improve training speed. In this paper, We optimize the trajectory in terms of time, which also has the effect of significantly accelerating training. An intuitive idea is to simply combine the above spatial optimization models with our approach. How to combine temporal and spatial optimization more effectively remains the focus of subsequent research.
\section{Conclusion}
We have presented TO-FLOW, a model that optimizes time and does not introduce additional computational cost. Excessive computational costs are a major bottleneck to scaling CNF to large applications. We integrate an additional temporal optimization step to the training process, which regularizes the trajectory from another perspective and significantly improves computational efficiency. Furthermore, our method is compatible with other regularization methods and can be applied to other more expressive architectures to further improve performance.

\noindent\textbf{Acknowledgements:} This work was supported in part by grants from National Science Foundation of China (61571005), the fundamental research program of Guangzhou, China (2020B1515310023).

\clearpage
{
    \clearpage
    \small
    \bibliographystyle{unsrt}
    \bibliography{macros,main}
}

\appendix

\setcounter{page}{1}

\twocolumn[
\centering
\Large
\textbf{TO-FLOW: Efficient Continuous Normalizing Flows with Temporal Optimization adjoint with Moving Speed} \\
\vspace{0.5em}Supplementary Material \\
\vspace{1.0em}
] 
\appendix

\section{Generic form of Temporal Optimization}
\label{App-A}

If optimizing $t_0$ and $T$ at the same time, then the objective function has the following form:
\begin{equation}
    \label{minimizet0}
  \begin{split}
&\mathop{\min}\limits_{\boldsymbol{\theta} \in \boldsymbol{\Theta}, t_0 \in \mathbb{R}, T \in \mathbb{R}} L(\boldsymbol{\theta},t_{0},T) \\
&= -\mathbb{E}_{p_{\mathbf{x}}}\{\log p(\mathbf{z}(t_{0}); \boldsymbol{\theta})\} \\
&= -\mathbb{E}_{p_{\mathbf{x}}}\left \{\log p(\mathbf{z}(T); \boldsymbol{\theta})  + \int_{t_0}^T Tr(\mathbf{J}(t, \boldsymbol{\theta}) )dt \right \},
\end{split}
\end{equation}

The optimization of $T$ has been given above, hence in this section we only need to show the derivation of $\frac{\partial L}{\partial t_0}$. Also, it's not easy to get the $\frac{\partial L}{\partial t_0}$ and $\log p(\mathbf{z}(t_0))$ are unknown, so we rewrite the formula and introduce the chain rule by using intermediate variable $\mathbf{z}(t_0)$ for simplicity. Then we get

\begin{equation}
\label{app:final}
\begin{split}
&\frac{\partial L(\boldsymbol{\theta}, t_{0}, T)}{\partial t_{0}} \\
&= -\frac{\partial \mathbb{E}_{p_{\mathbf{x}}}\{\log p(\mathbf{z}(t_0),\boldsymbol{\theta})\}}{\partial t_{0}} \\
&= -\frac{\partial \mathbb{E}_{p_{\mathbf{x}}}\{\log p(\mathbf{z}(T),\boldsymbol{\theta})\}}{\partial t_0}+ \mathbb{E}_{p_{\mathbf{x}}}\{ Tr(\mathbf{J}(t_{0}, \boldsymbol{\theta}))\}\\
&= -\mathbb{E}_{p_{\mathbf{x}}} \left \{ \frac{\partial \log p(\mathbf{z}(T),\boldsymbol{\theta})}{\partial \mathbf{z}(t_0)}\circ \frac{\partial \mathbf{z}(t_0)}{\partial t_0} \right \}+
\mathbb{E}_{p_{\mathbf{x}}}\{ Tr(\mathbf{J}(t_{0}, \boldsymbol{\theta}))\}.\\
\end{split}
\end{equation}
Once we get $\frac{\partial L}{\partial t_0}$, we put $(\frac{\partial L}{\partial t_0}, t_{0})$ into the temporal optimizer $Q$ as mentioned in the text and optimize $(t_{0}, T)$ together.



\section{Comparison of 2-dimensional images at different stages}
\label{App-B}
\begin{figure*}
\begin{center}

\includegraphics[width=0.99\textwidth]{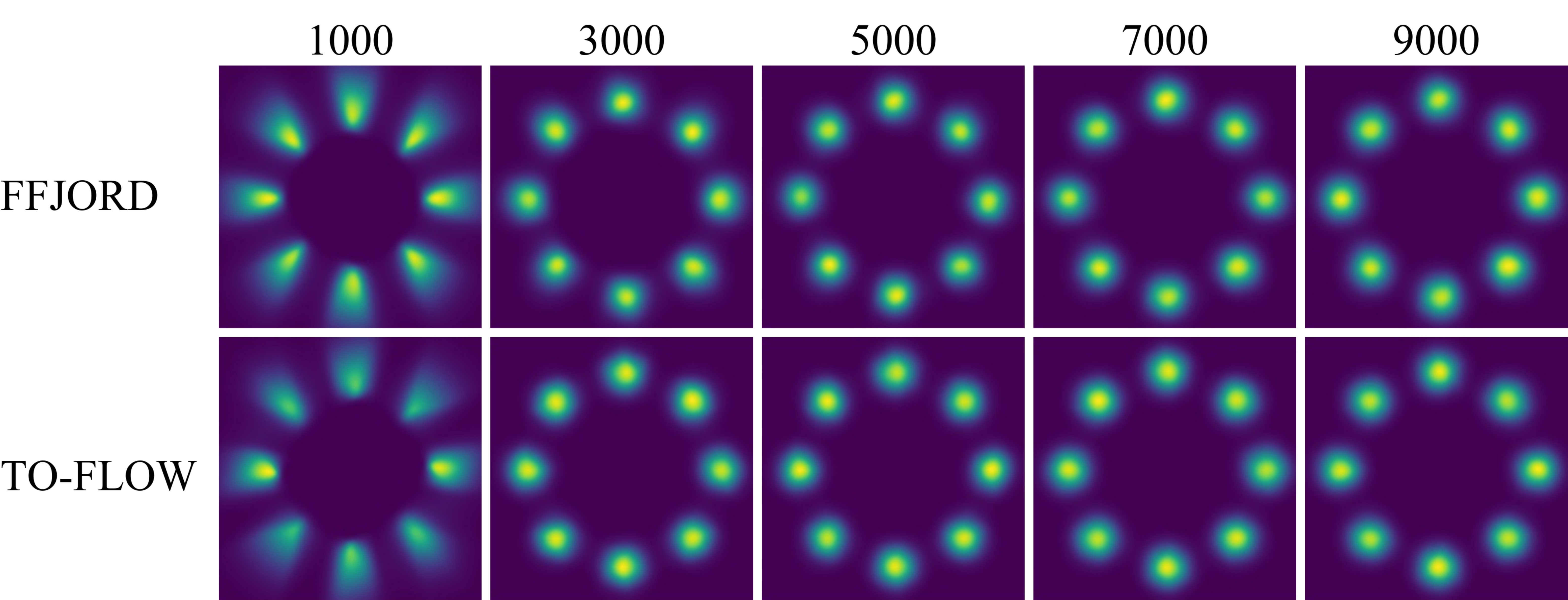}
\hfill
\end{center}
\caption{Comparison of FFJORD and TO-FLOW on 8gaussian data set. The numbers at the top of the images represent the number of iterations of the model.}
\label{fig:stage-8gaussian}
\end{figure*}

\begin{figure*}
\begin{center}

\includegraphics[width=0.99\textwidth]{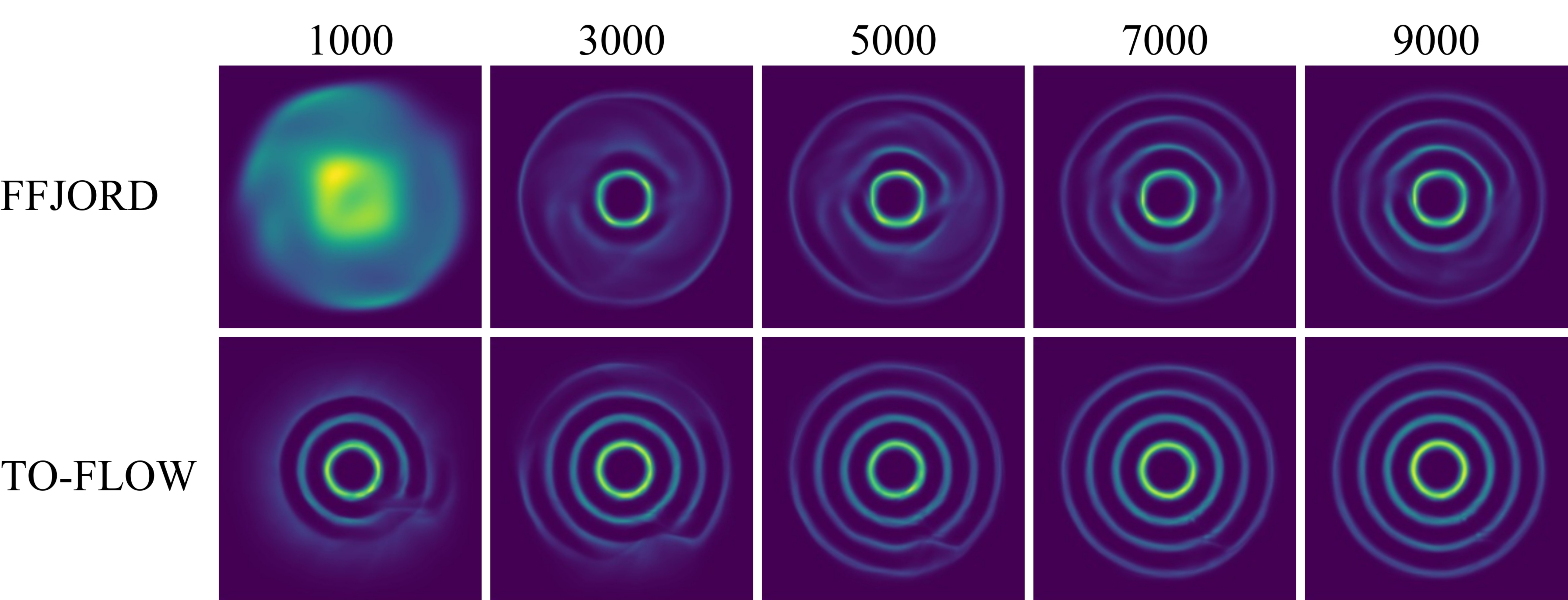}
\hfill
\end{center}
\caption{Comparison of FFJORD and TO-FLOW on rings data set. The numbers at the top of the images represent the number of iterations of the model.}
\label{fig:stage-rings}
\end{figure*}

\begin{figure*}
\begin{center}

\includegraphics[width=0.99\textwidth]{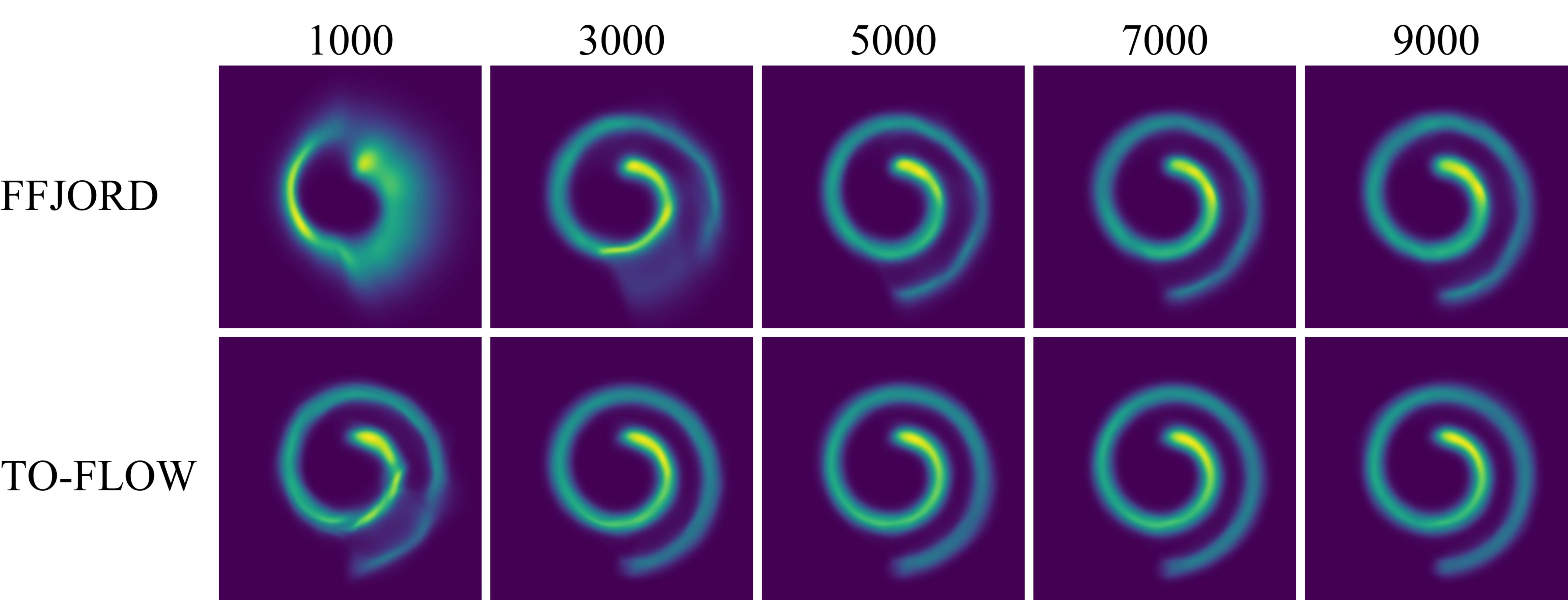}
\hfill
\end{center}
\caption{Comparison of FFJORD and TO-FLOW on swissroll data set. The numbers at the top of the images represent the number of iterations of the model.}
\label{fig:stage-swissroll}
\end{figure*}

\begin{figure*}
\begin{center}

\includegraphics[width=0.99\textwidth]{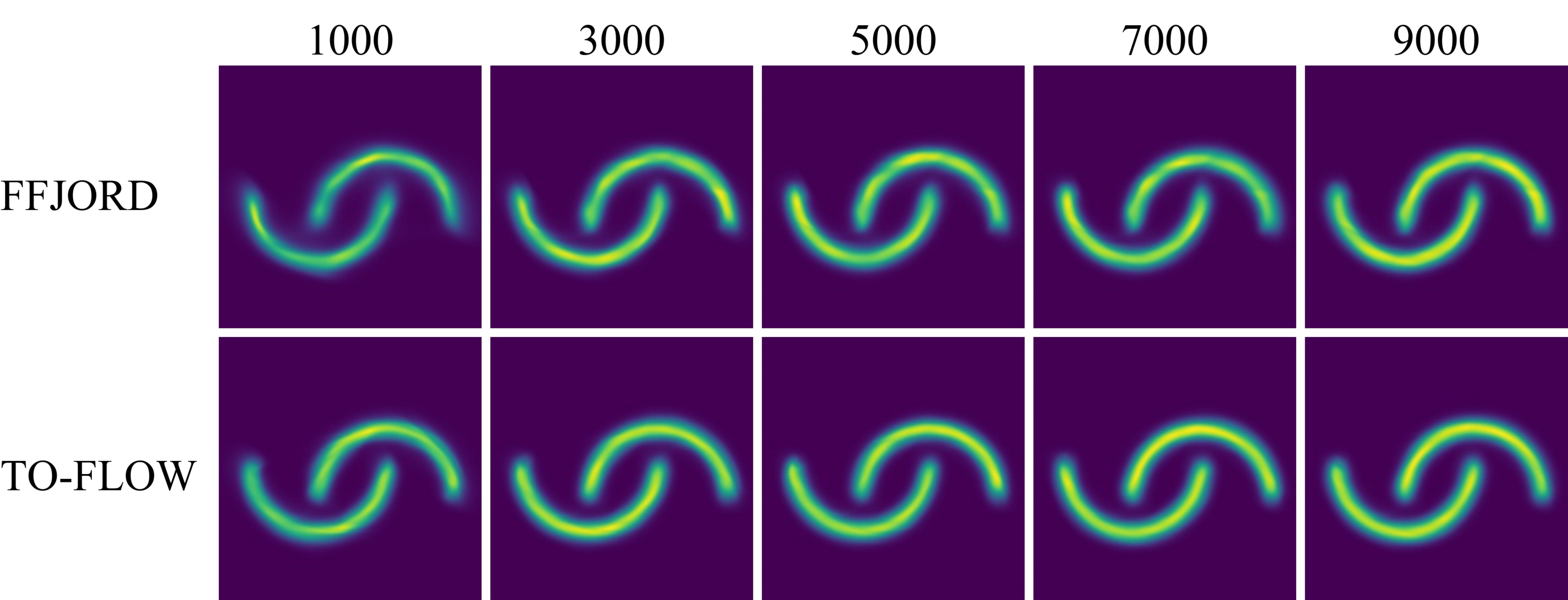}
\hfill
\end{center}
\caption{Comparison of FFJORD and TO-FLOW on moons data set. The numbers at the top of the images represent the number of iterations of the model.}
\label{fig:stage-moons}
\end{figure*}

\begin{figure*}
\begin{center}

\includegraphics[width=0.99\textwidth]{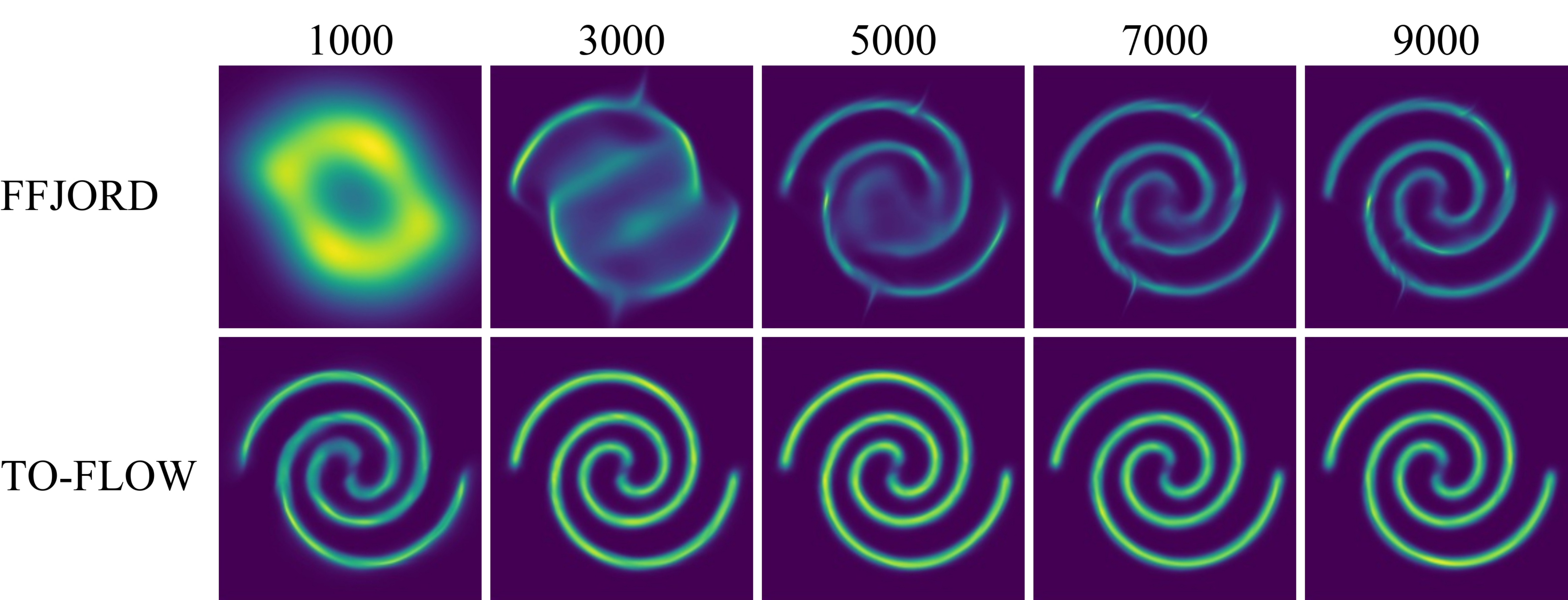}
\hfill
\end{center}
\caption{Comparison of FFJORD and TO-FLOW on 2spirals data set. The numbers at the top of the images represent the number of iterations of the model.}
\label{fig:stage-2spirals}
\end{figure*}

\begin{figure*}
\begin{center}

\includegraphics[width=0.99\textwidth]{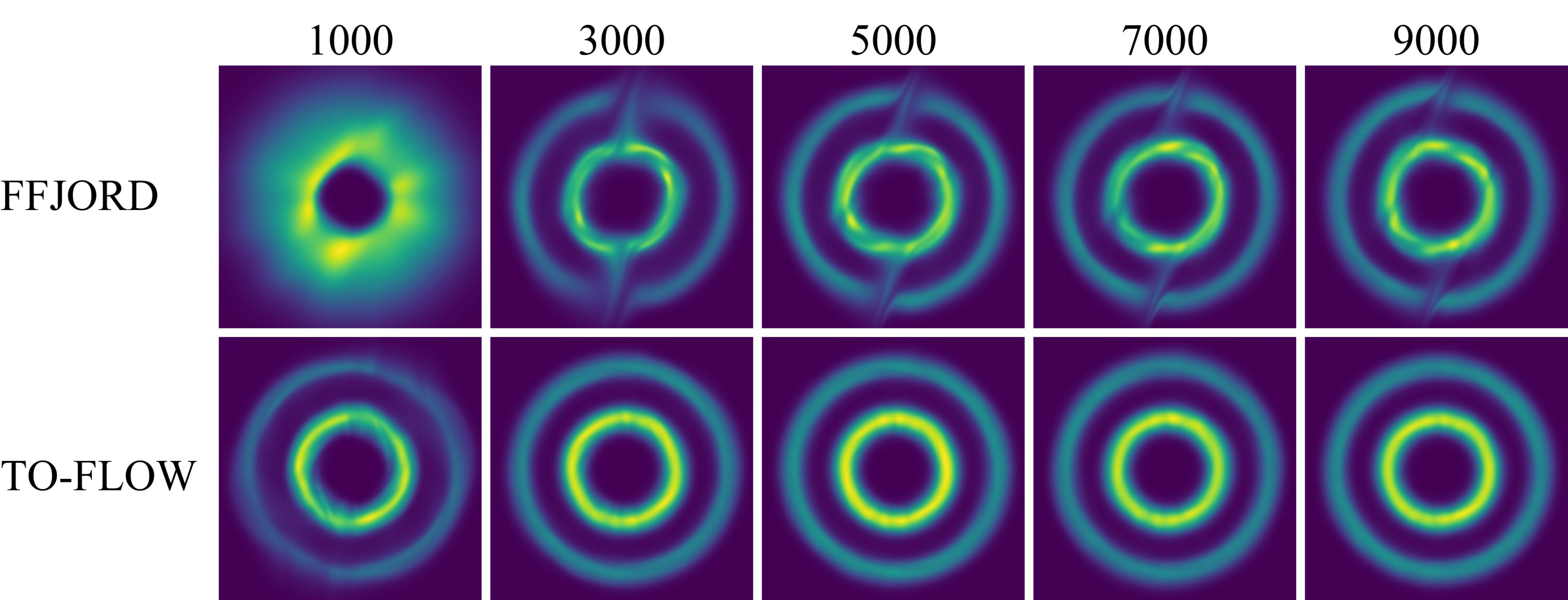}
\hfill
\end{center}
\caption{Comparison of FFJORD and TO-FLOW on circles data set. The numbers at the top of the images represent the number of iterations of the model.}
\label{fig:stage-circles}
\end{figure*}

\begin{figure*}
\begin{center}

\includegraphics[width=0.99\textwidth]{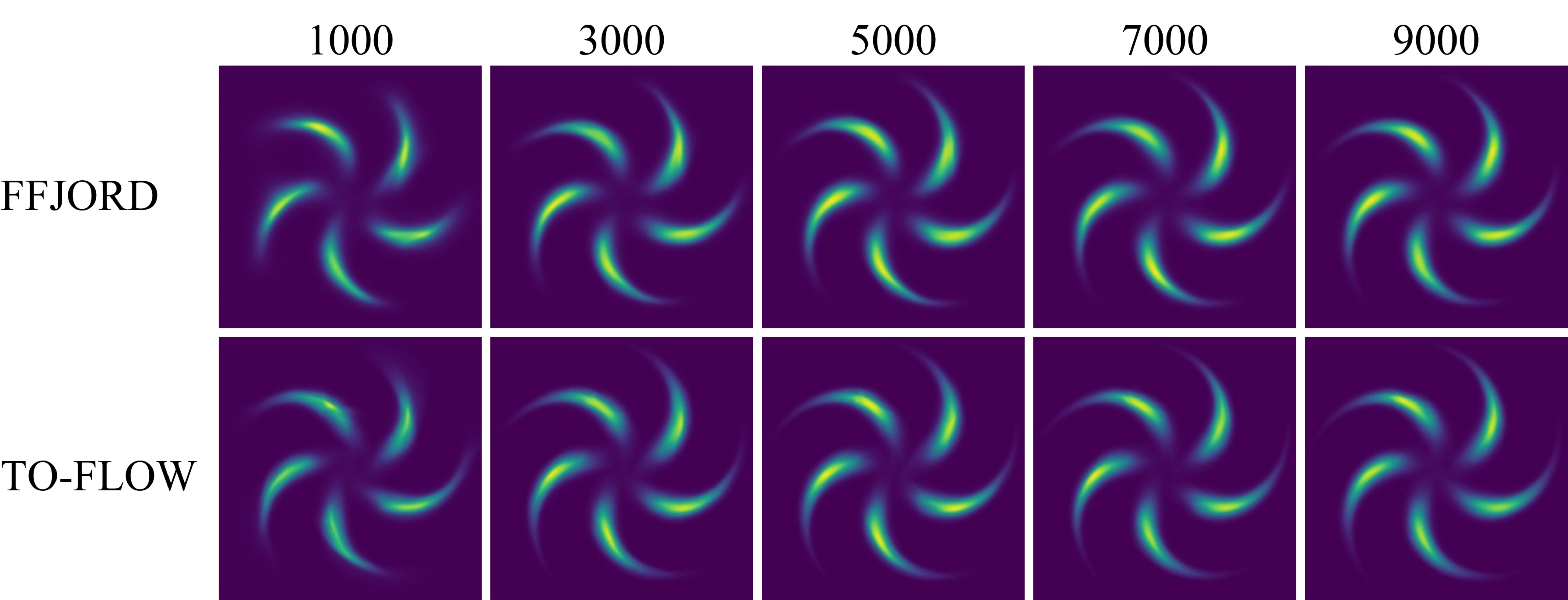}
\hfill
\end{center}
\caption{Comparison of FFJORD and TO-FLOW on pinwheel data set. The numbers at the top of the images represent the number of iterations of the model.}
\label{fig:stage-pinwheel}
\end{figure*}

We compare the generated images of FFJORD and TO-FLOW for eight 2D datasets at $1000$, $3000$, $5000$, $7000$ and $9000$ iterations. The results are shown in Figures  \ref{fig:stage-8gaussian},
\ref{fig:stage-rings},
\ref{fig:stage-swissroll},
\ref{fig:stage-moons},
\ref{fig:stage-2spirals},
\ref{fig:stage-circles} and
\ref{fig:stage-pinwheel}.

\section{Images under different temporal regularization}
\label{App-C}
\begin{figure*}
\begin{center}
\includegraphics[width=.99\columnwidth]{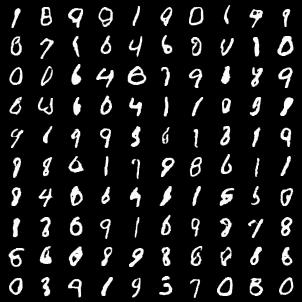}
\hfill
\includegraphics[width=.99\columnwidth]{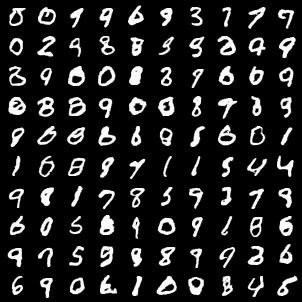}
\end{center}
\caption{Samples from MNIST. \\ 
Left: $\alpha = 0$ \quad Right: $\alpha = 0.1$}
\label{fig:option-mnist1}
\end{figure*}

\begin{figure*}
\begin{center}
\includegraphics[width=.99\columnwidth]{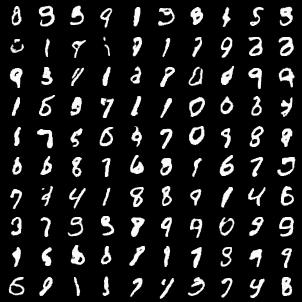}
\hfill
\includegraphics[width=.99\columnwidth]{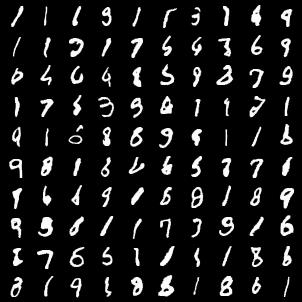}
\end{center}
\caption{Samples from MNIST. \\ 
Left: $\alpha = 0.2$ \quad Right: $\alpha = 0.3$}
\label{fig:option-mnist2}
\end{figure*}

\begin{figure*}
\begin{center}
\includegraphics[width=.99\columnwidth]{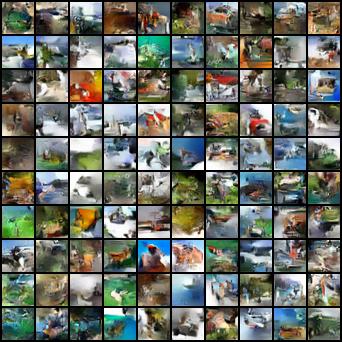}
\hfill
\includegraphics[width=.99\columnwidth]{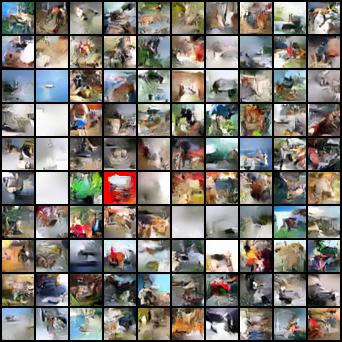}
\end{center}
\caption{Samples from CIFAR-10. \\ 
Left: $\alpha = 0$ \quad Right: $\alpha = 0.1$}
\label{fig:option-cifar1}
\end{figure*}

\begin{figure*}
\begin{center}
\includegraphics[width=.99\columnwidth]{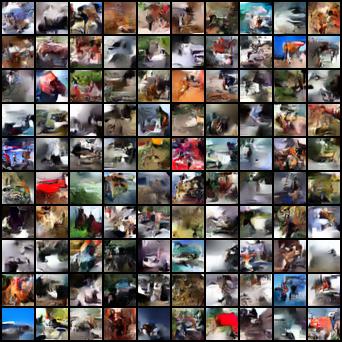}
\hfill
\includegraphics[width=.99\columnwidth]{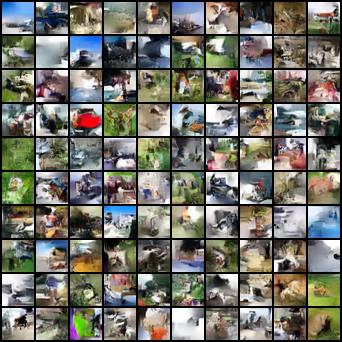}
\end{center}
\caption{Samples from CIFAR-10. \\ 
Left: $\alpha = 0.2$ \quad Right: $\alpha = 0.3$}
\label{fig:option-cifar2}
\end{figure*}

\begin{figure*}
\begin{center}
\includegraphics[width=.99\columnwidth]{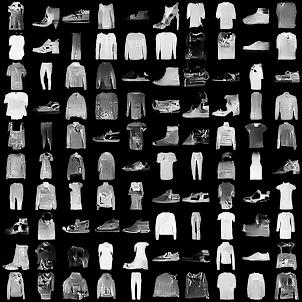}
\hfill
\includegraphics[width=.99\columnwidth]{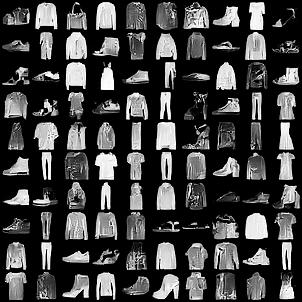}
\end{center}
\caption{Samples from FASHION-MNIST. \\ 
Left: $\alpha = 0$ \quad Right: $\alpha = 0.1$}
\label{fig:option-fashion1}
\end{figure*}

\begin{figure*}
\begin{center}
\includegraphics[width=.99\columnwidth]{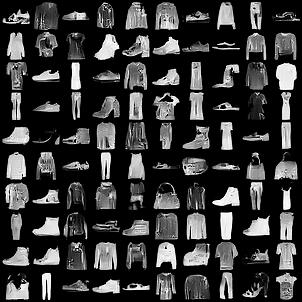}
\hfill
\includegraphics[width=.99\columnwidth]{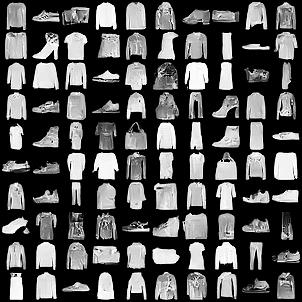}
\end{center}
\caption{Samples from FASHION-MNIST. \\ 
Left: $\alpha = 0.2$ \quad Right: $\alpha = 0.3$}
\label{fig:option-fashion2}
\end{figure*}

We generate images of the three datasets under different temporal regularization. The results are shown in Figures \ref{fig:option-mnist1}, 
\ref{fig:option-mnist2}, 
\ref{fig:option-cifar1},
\ref{fig:option-cifar2}, 
\ref{fig:option-fashion1}, 
\ref{fig:option-fashion2}.

\end{document}